\documentclass[manuscript,screen,review]{acmart}

\usepackage{algorithm}
\usepackage{algorithmic}
\usepackage{multirow}
\usepackage{url}
\AtBeginDocument{%
  }

\begin{document}

\title{Detection of Breast Cancer Lumpectomy Margin with SAM-incorporated Forward-Forward Contrastive Learning}

\author{Tyler Ward}
\email{tyler.ward@uky.edu}
\orcid{0000-0003-0669-1407}
\author{Xiaoqin Wang}
\email{xiaoqin.wang@uky.edu}
\author{Braxton McFarland}
\orcid{0009-0009-5069-2387}
\author{Md Atik Ahamed}
\email{atikahamed@uky.edu}
\orcid{0000-0002-7746-4247}
\author{Sahar Nozad}
\email{sahar.nozad@uky.edu}
\author{Talal Arshad}
\email{t.arshad@uky.edu}
\author{Hafsa Nebbache}
\email{hafsa.nebbache@uky.edu}
\affiliation{%
  \institution{University of Kentucky}
  \city{Lexington}
  \state{Kentucky}
  \country{USA}
}

\author{Jin Chen}
\email{jinchen@uab.edu}
\orcid{0000-0001-6721-3199}
\affiliation{%
  \institution{The University of Alabama at Birmingham}
  \city{Birmingham}
  \state{Alabama}
  \country{USA}
}

\author{Abdullah Imran}
\email{aimran@uky.edu}
\orcid{0000-0001-5215-339X}
\affiliation{%
  \institution{University of Kentucky}
  \city{Lexington}
  \state{Kentucky}
  \country{USA}
}

\renewcommand{\shortauthors}{Ward et al.}

\begin{abstract}
  Complete removal of cancer tumors with a negative specimen margin during lumpectomy is essential in reducing breast cancer recurrence. However, 2D specimen radiography (SR), the current method used to assess intraoperative specimen margin status, has limited accuracy, resulting in nearly a quarter of patients requiring additional surgery. To address this, we propose a novel deep learning framework combining the Segment Anything Model (SAM) with Forward-Forward Contrastive Learning (FFCL), a pre-training strategy leveraging both local and global contrastive learning for patch-level classification of SR images. After annotating SR images with regions of known maligancy, non-malignant tissue, and pathology-confirmed margins, we pre-train a ResNet-18 backbone with FFCL to classify margin status, then reconstruct coarse binary masks to prompt SAM for refined tumor margin segmentation. Our approach achieved an AUC of 0.8455 for margin classification and segmented margins with a 27.4\% improvement in Dice similarity over baseline models, while reducing inference time to 47 milliseconds per image. These results demonstrate that FFCL-SAM significantly enhances both the speed and accuracy of intraoperative margin assessment, with strong potential to reduce re-excision rates and improve surgical outcomes in breast cancer treatment. Our code is available at ~\url{https://github.com/tbwa233/FFCL-SAM/}.

\end{abstract}

\maketitle

\section{Introduction}
\label{sec:introduction}
Breast-conserving surgery (BCS), or lumpectomy, is a standard surgical treatment for women with early-stage breast cancers \cite{siegel2023cancer, katipamula2009trends}. Annually, \textit{over 200,000 lumpectomies} are performed in the United States \cite{lautner2015disparities}. Detection of breast cancer on margins during breast-conserving surgery (also known as lumpectomy), a surgical procedure to remove tumors or cancer-affected portions of the breast, is crucial to reduce the risk of cancer recurrence \cite{ho2020deep, d2021multi}. Accurate assessment of margins ensures the complete removal of cancerous tissue, avoiding the need for re-excision and re-operation \cite{pradipta2020emerging}. While lumpectomy has served as the primary treatment for early-stage breast cancer, there is a need for more precise techniques in evaluating resection margins towards more enhanced surgical outcomes.

\begin{figure}
    \centering
    \includegraphics[width=\linewidth]{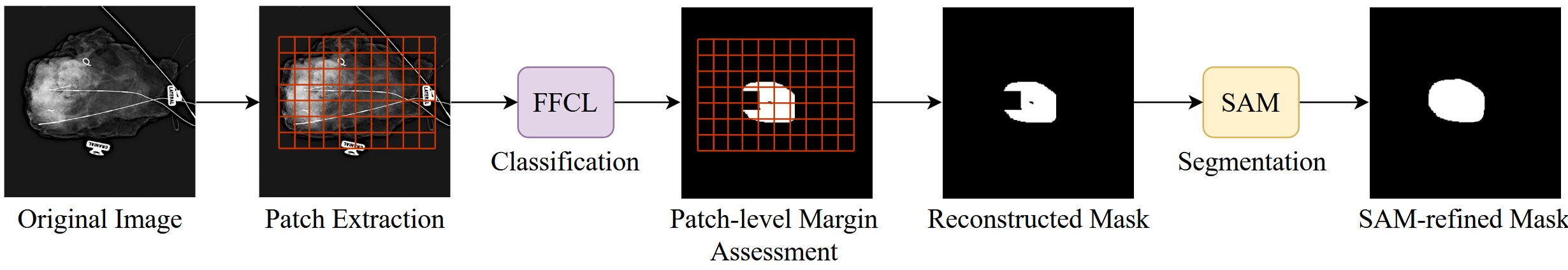}
    \caption{Overview of the proposed FFCL-SAM workflow: First, FFCL is employed to predict margin status at the patch-level on an SR image. After reconstructing the coarse margin mask, SAM-based refinement is performed to obtain the final margin masks.}
    \label{fig:FFCL-overview}
\end{figure}

Radiography-based inspection, the standard practice during BCS for margin assessment, has limited power in identifying tumor margins. Intraoperative digital specimen radiograph (SR), a form of X-ray imaging, has been routinely employed by radiologists to assess cancer removal and specimen margins during BCS \cite{chen2023analysis}. In X-ray imaging, the visual distinction between soft tissues of similar density, such as tumors and other structures in the breast, is often minimal. While lumpectomy has been the main treatment for early-stage breast cancer, more precise margin evaluation is needed for better surgical outcomes. SR has relatively \textit{low sensitivity, ranging between 36-58\%}~\cite{scimone2021assessment}, resulting in the need for subsequent surgeries in approximately \textit{one in four patients} \cite{scimone2021assessment}. SR-based margin assessment struggles primarily due to the challenge of differentiating soft tissues of similar density, e.g., tumors from surrounding breast structures. Such ineffectiveness can further lead to increased psychological and physical burdens for those undergoing treatment. Incorporating artificial intelligence (AI) into an intraoperative setting could be a better alternative in assessing surgical margins and ensuring complete tumor resection. 

With the advent of AI, specifically deep learning techniques, medical image analysis can be performed more effectively. Unlike other medical imaging applications, deep learning-based margin assessment on SRs is still underexplored. Existing AI-based margin assessment methods are primarily supervised, requiring human-labeled data for model training~\cite{d2021multi, veluponnar2023toward,david2023situ, chen2023analysis, to2022deep, lu2022automated}. Labeling SR images involves reading the corresponding pathology reports and manually segmenting margins, a notably labor-intensive process. Therefore, it is critical for improved and automated intraoperative image analysis techniques that can precisely assess surgical margins while lowering manual labor \cite{garza2024intraoperative}. 

Recently, vision foundation models for various image analysis tasks have gained a lot of attention due to their impressive generalizability and adaptability to limited data settings~\cite{awais2025foundation}. However, these models need to be better adapted to margin assessment, as they are primarily developed on nonsurgical data. Surgical scenes have significantly different characteristics compared to natural ones~\cite{ma2021comprehensive}, such as varying textures, lighting conditions, and anatomical structures. Adapting these models specifically for margin assessment would enhance their ability to \textit{accurately identify and delineate} cancer margins in complex and variable surgical environments, leading to \textit{more precise and reliable} intraoperative image analysis. 

This work proposes FFCL-SAM by integrating the Segment Anything Model (SAM) with the Forward-Forward Contrastive Learning (FFCL)~\cite{ahamed2023ffcl}. FFCL employs the Forward-Forward Algorithm (FFA)~\cite{hinton2022forward} and performs two-stage pre-training (local followed by global contrastive learning) without requiring any backpropogation. FFCL has been successfully used for pneumonia detection from chest X-rays~\cite{ahamed2023ffcl} and BCS margin assessment~\cite{ahamed2024automatic, ward2025automated}. SAM is a promptable segmentation that can be used for medical image segmentation~\cite{kirillov2023segment}. Prior SAM-based medical image segmentation works are primarily reliant on manual or semi-automated prompting for accurate segmentation~\cite{zhou2024sam}. SAM can be vulnerable and may perform poorly when attempting an automated prompting without much domain knowledge. Considering the effectiveness of FFCL in assessing margin assessment at the patch level~\cite{ahamed2024automatic, ward2025automated}, we can exploit the domain-specific pre-training to guide SAM for improved segmentation of positive margins on SRs (Fig.~\ref{fig:FFCL-overview}). To our best knowledge, the SAM model, as well as integration with an FFA-based approach (FFCL), have not been integrated for medical image analysis before, let alone the intraoperative margin assessment. 

Deviating from the existing margin assessment techniques, we, therefore, develop a novel SAM-incorporated positive margin assessment method that leverages FFCL for initial mask generation detecting margins at the patch level. The final positive margin is detected by refining the initial mask by SAM.       

Identifying visually imperceptible indicators of positive lumpectomy margins could potentially reduce cancer recurrence and re-excision rates of breast cancer patients. Our contributions in the presented paper can be summarized as:
\begin{itemize}
    \item Novel SAM-incorporated efficient and precise BCS margin assessment from intraoperative specimen radiographs through pixel-wise labeling;
    
    \item Image-level classification of positive margins from specimen radiographs leveraging an effective pretraining strategy FFCL;

    \item Final segmentation of positive margins using a promptable SAM-based segmentation refinement;

    \item Extensive evaluation demonstrating the effectiveness of the proposed FFCL-SAM in accurately detecting lumpectomy margins on SR images.
\end{itemize}

\paragraph*{Our Previous Work:}
This manuscript is an extension of our paper ``Automatic detection of breast cancer lumpectomy margin from intraoperative specimen mammograms'' presented at the International Workshop on Breast Imaging (IWBI 2024)~\cite{ahamed2024automatic}. This manuscript substantially extends by adding a thorough literature review, experiments with larger datasets, additional experiments and baselines, improved results, a more detailed description of the methods and results discussion, and additional figures and visualizations. The largest addition comes from our modified architecture. 

Our IWBI paper primarily focused on patch-level classification for patches extracted from intraoperative specimen mammography radiographs.  We employed the ResNet-18 model pre-trained using our innovative FFCL strategy. In this manuscript, we introduce a reconstruction method into our detection pipeline to produce rough segmentation masks based on the binary predictions of the classification model. We then employ a promptable segmentation model to refine these masks.

\section{Related Work}
\label{sec:related-work}

\subsection{Self-Supervised Representation Learning}
In deep learning, the process of attempting to extract more abstract and useful information from an input by learning a parametric mapping from the raw data domain to a feature vector or tensor is called representation learning\cite{lekhac2020contrastive}. As with many deep learning models, the main barrier preventing the widespread practical deployment of representation learning models lies in their reliance on large labeled datasets, which often carry a high annotation cost. In recent years, self-supervised representation learning (SSRL) has emerged as a method to reduce this bottleneck. Unlike fully supervised models, SSRL models employ carefully designed pretext tasks to discriminatively train deep representations, which enable the solution of downstream tasks with much smaller amounts of task-specific labeled data compared to fully supervised methods\cite{ericsson2022self}.

Contrastive learning is a form of SSRL that aims to train a model train a model to differentiate between similar and dissimilar samples\cite{hu2024a} Seminal work like SimCLR\cite{chen2020a}, MoCo\cite{he2020momentum}, and BYOL\cite{grill2020bootstrap} have established contrastive learning as a highly effective approach for generating high-quality embeddings, particularly in imaging tasks. In such tasks, a similar (positive) pair consists of different views of the same image, while a dissimilar (negative) pair includes representations from separate images. Through this process, the contrastive learning model learns to group similar features together while distinguishing different data points, resulting in embeddings that are more discriminative and relevant for the specific task.

In medical imaging, contrastive learning and other SSRL techniques have shown significant promise. For example, contrastive pre-training has demonstrated effectiveness in improving the accuracy, generalizability, and efficiency of dermatology condition classification from digital camera images\cite{azizi2021big}. Similar applications have been seen in radiography\cite{huang2024enhancing} and mammography\cite{li2021domain}, where contrastive learning has enabled better feature extraction with fewer annotations. 

Despite these promising applications, many SSRL models face limitations in terms of actual deployment in real-time medical settings\cite{ye2024continual}. Recent research has highlighted the fact that one of the core components of modern deep learning models, backpropagation, can lead to high power consumption and intense resource allocation\cite{muller2024resource}. As such, a new focus within medical imaging research has been the development of backpropagation-free models, employing methods such as the forward-forward algorithm\cite{hinton2022forward}. This method has been used to create classification models for X-ray\cite{ahamed2023ffcl}, dermatology\cite{angulo2023the}, and mammography\cite{ahamed2024automatic, ward2025automated} data, with great success, highlighting its potential for reducing the dependence on backpropagation in deep learning-based medical imaging applications.

\subsection{Medical Image Segmentation}
While SSRL largely revolves around pre-training models that are large, unlabeled datasets, segmentation models are used to delineate regions of interest, such as tumor margins, from a scan at the pixel level. For years, encoder-decoder architectures inspired by U-Net\cite{ronneberger2015u} have served as the standard starting point for medical image segmentation models \cite{azad2024medical}. While such structures are undoubtedly powerful, particularly in their ability to learn spatial hierarchies and recover fine-grained details, they do suffer from several of the same drawbacks as other deep learning architectures, namely the vanishing gradient problem \cite{pribadi2024optimization}, where as an architecture deepens, the gradients that are used to update the neural network tend to become smaller and smaller until they "vanish" as they are backpropagated through the layers of the network. A popular method of addressing this problem is to incorporate residual connections into the U-Net architecture \cite{li2024mresunet}. This approach has been shown to improve training efficiency and allow for deeper models by affording better gradient flow across layers\cite{he2016deep}. However, while this method does increase performance on complex segmentation tasks, it can still struggle with capturing global context in large medical images, particularly when dealing with long-range dependencies \cite{pramanik2024daunet}.

More recent medical image segmentation architectures, such as TransUNet\cite{chen2021transunet, chen2024transunet} and SwinU-Net\cite{cao2022swin}, have emerged as strong alternatives to these earlier methods. TransUNet incorporates transformer layers into the U-Net encoder to extract global contexts from images, while the decoder upsamples the encoded features and combines them with high-resolution feature maps to enable precise localization\cite{chen2021transunet}. This architecture is particularly useful for tasks where the spatial relationships between regions can hold critical diagnostic information, such as organ boundary delineation or lesion segmentation\cite{chen2024transunet}. Similarly, SwinU-Net builds upon the Swin Transformer\cite{liu2021swin} framework, introducing a hierarchical attention mechanism that can efficiently model both global and local features while maintaining computation efficiency. It also uses a window-based attention scheme for better scalability to larger medical images. Despite the power of these transformer-based medical image segmentation models, they are limited to high computational complexity and increased resource allocation.

Promptable segmentation models such as the Segment Anything Model (SAM) \cite{kirillov2023segment} offer a promising avenue for avoiding the high computational requirements of other transformer-based methods due to their strong zero-shot and few-shot generalizability. SAM supports a variety of different prompt formats, including points \cite{xu2023sppnet}, bounding boxes \cite{rahman2024pp}, text \cite{zhang2024evf}, and masks \cite{xie2024masksam}.  The latest version of SAM, SAM 2 \cite{ravi2024sam}, is capable of segmenting videos as well as images. However, manual/semi-automated prompting for SAM carries several of the same flaws as the manual annotation of training data in that they can be time-consuming to generate. A potential solution for the issues present in manual or semi-automated prompting of SAM lies in the incorporation of SSRL techniques. 

\subsection{Breast Cancer Margin Assessment}
Achieving negative margins during breast-conserving surgeries like lumpectomy is crucial for reducing the risk of local recurrence in breast cancer patients. A negative margin, defined as the absence of tumor cells at the edges of the resected tissue, ensures complete tumor removal while preserving healthy tissue\cite{rakha2024revisiting}. Despite advances in surgical techniques and imaging, approximately 20\% of patients require re-excision due to positive margins, leading to increased healthcare costs, worse cosmetic outcomes, and added psychological stress for patients\cite{chakedis2022economic}.

Traditional methods for assessing lumpectomy margins include frozen section analysis, imprint cytology, and specimen radiography. While frozen section analysis provides high diagnostic accuracy, it is time-consuming and requires specialized resources that may not be available in all surgical centers \cite{ahuja2024comparison}. Imprint cytology is faster but less reliable, particularly for identifying ductal carcinoma in situ (DCIS) and subtle invasive components \cite{esbona2012intraoperative}. Specimen radiography, which involves 2D imaging of the excised tissue, is the most commonly used intraoperative method due to its ease of use and widespread availability \cite{dowling2024diagnostic}. However, radiography has limited sensitivity (36–58\%) for detecting positive margins, especially for lesions with low contrast or diffuse growth patterns \cite{scimone2021assessment}. 

The limitations of these traditional methods have opened the door for the utilization of more advanced imaging and computational techniques, particularly those guided by AI methods. Recent advances in deep learning have demonstrated the potential for improving breast cancer margin detection, although they are not without limitations. For example, one study \cite{lu2022automated} incorporated an automated margin detection based on the texture analysis of microscopy with ultraviolet surface excitation (MUSE) images using a support vector machine (SVM)-based two-stage classification approach. Another study \cite{david2023situ} also employed SVM-based classification, this time on hand-crafted features obtained from Raman spectroscopy. Despite generating positive results, both of these methods come with the very large drawback of relying on largely human-driven annotation and analysis methods, limiting their scalability and generalizability to other tasks.

Several studies have explored the AI-driven techniques for breast cancer detection and margin assessment in whole slide images (WSIs). To address concerns regarding small dataset sizes when working with WSIs for this purpose, a common approach is to extract patches from the WSIs and use those to initialize model training. One study \cite{ho2020deep} sought to improve AI models' understanding of regional importance to this type of classification by fusing explainable features generated from a ResNet classifier trained on patches extracted from whole slide deep ultraviolet (DUV) imaged using a decision method, finding that incorporating these features improved the results compared to the baseline ResNet method. Another study sought to evaluate AI's potential as a screening tool for the pathological assessment of margins, training a deep convolutional neural network (CNN). Both of these studies demonstrated positive results in terms of the problems they set out to solve, but neither fully addressed the challenge of poor margin localization, and it was found, particularly in the latter study, that these patch extraction approaches can lead to misclassifications if critical cancerous regions are small or fragmented across multiple patches.

Other studies have explored the general applicability of a wide spectrum of AI-driven models for margin assessment. One study \cite{chen2023analysis} evaluated the performance of four difference classifiers (ResNet-50, InceptionV3, Inception-ResNet-v2, and DenseNet-121) and two different pre-training techniques (RadImageNet and ImageNet), finding that InceptionV3 pre-trained on RadImageNet offered the best performance in predicting the pathologic margin status of breast cancer specimens using specimen mammography images. Combining the information obtained from multiple classifiers in an ensemble has also proven effective, as shown in a study \cite{veluponnar2023toward} where eight pre-trained networks (AlexNet, U-Net, VGG16, VGG19, ResNet18, ResNet50, MobileNet-V2, and Xception) were ensembled to achieve improved results compared to the individual performance of these methods. However, the ensemble approach carries the drawback of increasing the resource intensity of running the models, limiting the scalability of this approach in practice. In summary, there is great potential for AI-driven methods for margin assessment, but there are still gaps and challenges that need to be explored.

\section{Methods}
\label{sec:methods}

\subsection{Problem Formulation}
Given a set of lumpectomy radiograph images $X$, our goal is to learn a mapping function $f: X \to Y$ that predicts the corresponding segmentation mask $Y$, where each pixel is labeled as either belonging to a positive margin or not. Instead of trying to learn this mapping directly, we first define an intermediate classification task at the patch label. Given $x_i \in x$, we partition into a set of overlapping patches $P = \{p_1, p_2, ..., p_n\}$, where each patch $p_k$ is associated with a binary label $y_k \in \{0,1\}$ indicating the presence of absence of a positive margin.

For classification, we employ FFCL, an SSL approach that enhances feature representation through a two-stage pretraining strategy: local contrastive learning within individual layers and global contrastive learning across the entire network. The FFCL-pretrained model is then fine-tuined on labeled patches to perform binary classification. The predicted labels from the classified patches are aggegrated and mapped back to their corresponding spatial location in the original radiograph to reconstruct a coarse binary mask $M_c$, where $M_c(x, y) = 1$ if the coresponding patch was classified as positive, and $M_c(x, y) = 0$ otherwise.

To refine the coarse mask, we employ SAM. Specifically, two prompts are generated from $M_c$, a bounding box $p_1$ that encloses the coarse mask region and a mask prompt $p_2$ based on $M_c$ itself. SAM first generates an initial mask $m_1$ using $p_1$ and further refines it into the final segmentation output $m_2$ using $p_2$.

\subsection{Preliminaries}

\paragraph{Forward-Forward Algorithm (FFA):}
FFA\cite{hinton2022forward} seeks to address several of the challenges with backpropagation discussed in the preceding sections. To compute the correct derivatives during the forward pass of a neural network, perfect knowledge of the computations performed is required. If a black box is inserted into the forward pass, backpropagation becomes impossible unless a differentiable model of the black box is learned. Reinforcement learning can be a solution here, but reinforcement learning approaches suffer from high variance and scale poorly, resulting in them being unable to compete with backpropagation in networks with potentially billions of parameters.

FFA employs a goodness function to facilitate the replacement of the forward and backward passes of backpropagation with double forward passes, one positive and one negative. The goal of the positive pass is to increase goodness in every hidden layer, while the negative pass does the opposite. The goodness function is defined as

\begin{equation}
P(pos) = \sigma(G - \theta),
\end{equation}

\noindent where  $\sigma$ is a logistic function, and $\theta$ is some threshold. The actual goodness calculation, $G$, is mathematically expressed by

\begin{equation}
G = \sum_{u} h_u^2.
\end{equation}

\begin{figure}
    \centering
    \includegraphics[width=\linewidth]{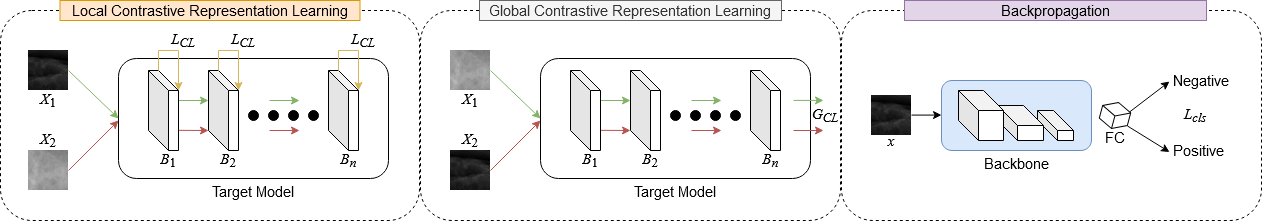}
    \caption{The proposed Forward-Forward Contrastive Learning (FFCL) for breast cancer margin detection. FFCL training involves multiple stages: first, a backbone architecture is selected as the target model. FFCL then trains the target model with local contrastive learning at each layer (Forward-Forward part), followed by global contrastive learning at the whole network (regular contrastive learning). Finally, the pre-trained model is used to perform the downstream classification task (positive/negative margins) using backpropagation.}
    \label{fig:ffcl}
\end{figure}

\noindent where $h$ is the activity of hidden unit $u$ before layer normalization.

\paragraph{Forward-Forward Contrastive Learning (FFCL):}
We adopt the FFCL-based pre-training strategy \cite{ahamed2023ffcl}. FFCL comprises two stages of pretraining before performing regular back-propagation for the downstream classification tasks. Fig.~\ref{fig:ffcl} shows the FFCL framework for positive margin detection. In the first stage, we perform contrastive learning locally based on the modified Forward-Forward Algorithm (FFA)~\cite{hinton2022forward}. FFCL is fed by two images $x_1, x_2 \in X_{train}$ sampled from the train set ($X_{train}$) and provides embeddings $E_{{x_1}_i}, E_{{x_2}_i}$ from each block $B_i$ followed by a ReLU activation. Global contrastive learning also takes two random images as input and maps to the final embedding space ($E_{x_1}, E_{x_2}$). Finally, the pre-trained model is used with regular backpropagation for performing the actual downstream margin detection.

\begin{figure}
    \centering
    \includegraphics[width=\linewidth]{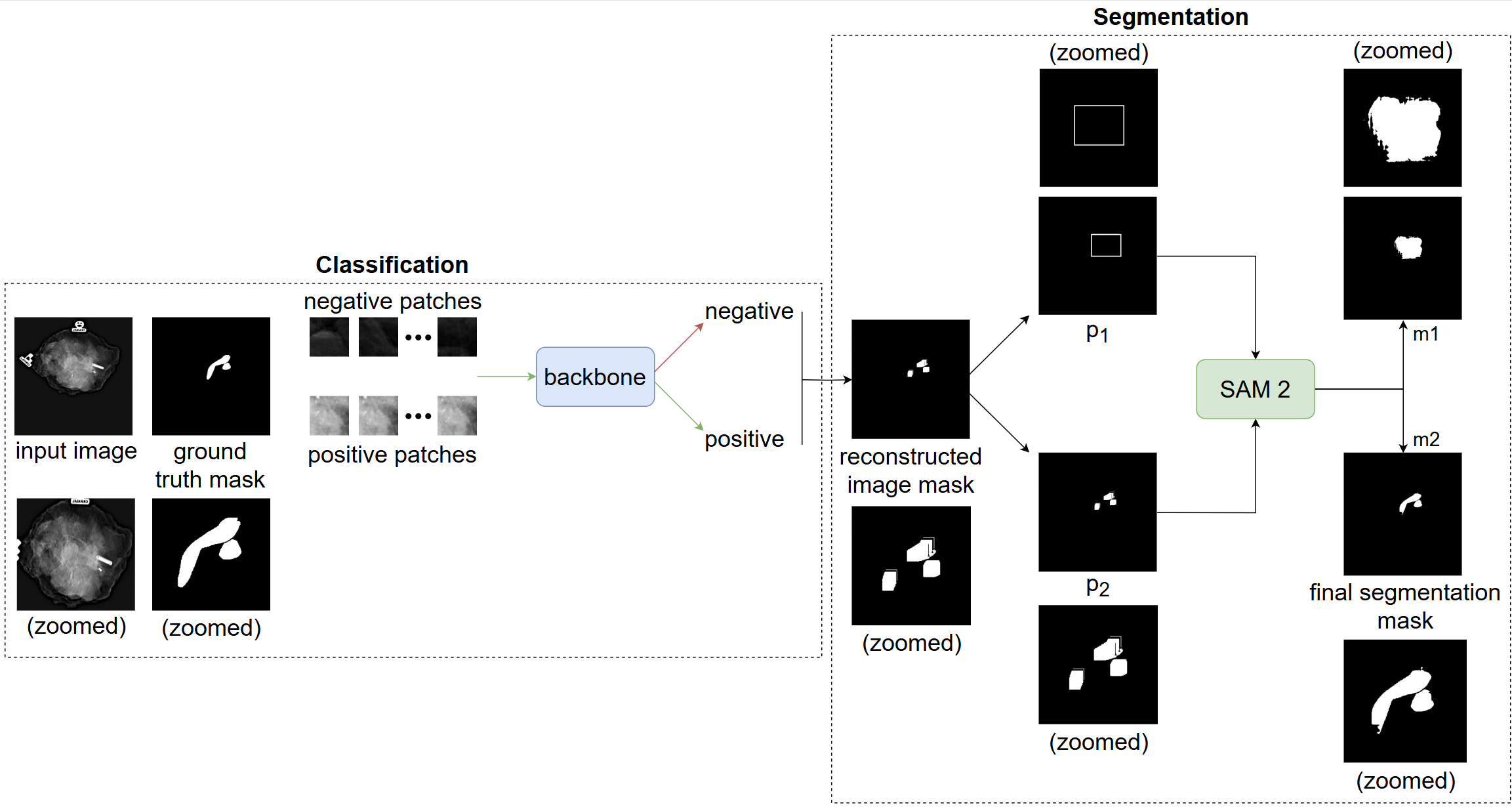}
    \caption{Detailed illustration of the breast cancer margin detection pipeline consisting of both classification and segmentation tasks. Each of the patient radiographs is annotated with pixel-level positive and negative tumors. Patches extracted from the segmented regions, are used to train the FFCL-pretrained model to detect positive margins. A rough binary mask is then reconstructed leveraging the class label predictions. Two prompts, $p_1$, which is a bounding box encompassing the coarse mask data, and $p_2$, which is the coarse mask itself, are generated from the coarse mask and used as input to the SAM 2 model. $p_1$ is used to produce the first mask prediction, $m_1$, and $p_2$ is used to refine the mask into the final segmentation mask, $m_2$.}
    \vspace{4pt}
    \label{fig:ffclsam}
\end{figure}

\subsection{FFCL-SAM}
We propose a two-stage approach, FFCL-SAM, aimed at improving the detection and segmentation of positive margins in intraoperative lumpectomy radiographs. The first stage focuses on patch-based classification using an FFCL-pretrained model. After classification labels have been obtained for the patches, they are reconstructed to form a coarse binary mask. The second stage of our proposed approach refines this segmentation mask by prompting SAM to produce more pixel-accurate segmentation.

\subsubsection{Model Design}
The full pipeline for FFCL is illustrated in Fig.~\ref{fig:ffclsam}. In the first stage, the FFCL-pretrained classification model is used to classify extracted 64x64 patches as either positive or negative, where positive means they contain pathology-confirmed positive margin, and negative means the patch contains only non-malignant tissue. The FFCL pretraining process is in itself a two-step process. In the first stage of pretraining, local contrastive learning is applied at each layer of the model to refine embeddings, while global contrastive learning is performed in the second stage to optimize the representations across the entire network. 

Once the FFCL pretraining is complete, the pre-trained model is then fine-tuned on fully labeled patches. A binary classification label for each patch is produced from this fine-tuning process, and the patches are then mapped back to their original spatial position in the radiograph using the inverse of the method used to extract them in the first place. This re-mapping process results in a coarse binary mask, where white regions correspond to patches that were predicted as positive, and black regions correspond to patches that were predicted to be negative. Prompts are then generated based on the coarse mask and are passed to SAM based on the coarse binary mask to refine the coarse mask into a refined mask that is closer to the ground truth.

\subsubsection{Patch Extraction}
We extract patches from the positive (center) tumor, negative, and positive margin regions. Using
the segmentation masks provided by the human annotator, patch extraction is performed by setting
the patch size to 64 × 64. Considering the imbalanced positive and negative regions, we extracted overlapping patches with stride = 3 for the positive class. For the negative class, the stride was set to 45, 35, and 60 for train, validation, and test, respectively.

\begin{figure}
    \centering
    \includegraphics[width=\linewidth]{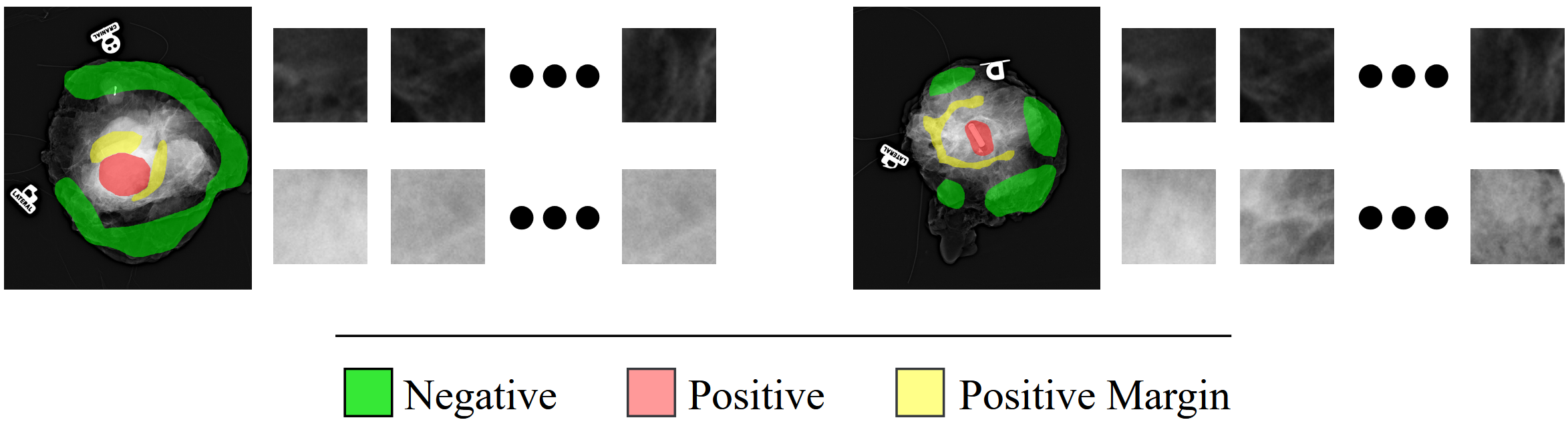}
    \caption{Sample patch extraction process from positive and negative margin regions of two different SR images.}
    \vspace{4pt}
    \label{fig:patchextraction}
\end{figure}

\subsubsection{Training Objectives}
During training, the primary goal of the classification stage is to train the FFCL-pretrained model to effectively distinguish between positive and negative patches. The strength of the FFCL pretraining technique is that is emphasizes the creation of discriminative feature embeddings via contrastive learning. In the first step of FFCL, the model learns to differentiate between embeddings at each layer, grouping similar features together while separating dissimilar ones. The second step of FFCL then focuses on refining the embeddings from the previous step across the entire network, with the goal of improving consistency and robustness. Both the local and global pretraining stages minimize the same cosine embedding loss function:

\begin{equation}
\operatorname{L_c}(E_{x_1},E_{x_2},C_{x_1},C_{x_2})= \begin{cases}1-\frac{E_{x_1} \cdot E_{x_2}}{\lVert E_{x_1} \rVert_2 \lVert E_{x_2} \rVert_2}, & \text { if } C_{x_1}=C_{x_2} \\ \max \left(0, \frac{E_{x_1} \cdot E_{x_2}}{\lVert E_{x_1} \rVert_2 \lVert E_{x_2} \rVert_2}\right), & \text { if }  C_{x_1}\neq C_{x_2}\end{cases}
\label{eq:loss_contrastive}
\end{equation}

\noindent where,  $(E_{x_1}, C_{x_1}) \ \&\ (E_{x_2}, C_{x_2})$ denote the embedding space and class label of the corresponding input images.

After pretraining, the model is then fine-tuned using a supervised approach where the model is trained to minimize a loss function to maximize classification accuracy. Considering the high imbalance in our dataset, we leverage the Focal cross-entropy loss \cite{lin2017focal} that helps the model to focus more on the minority class. For model-predicted classification probability $p$, the loss function is, therefore, defined as:
\begin{equation}
   L_(p) =  -\alpha_i(1–p)^\gamma log(p),
\end{equation}
where, $\alpha$ is a weighting parameter and $\gamma$ is a tunable focusing parameter. 

\begin{algorithm}[t]
\caption{SAM-based Mask Refinement}
\label{alg:FFCL-SAM}
\begin{algorithmic}
\REQUIRE 
    \STATE Image $I$
    \STATE Coarse segmentation mask $M_c$
    \STATE SAM2 predictor $S$
    \STATE Temporary video directory $D_{temp}$
\REPEAT
    \STATE Create single-frame video format: $V \leftarrow \{I\} \in D_{temp}$
    \STATE Initialize SAM2 state: $state \leftarrow S(V)$

    \STATE Extract bounding box: $p_1 \leftarrow \text{ComputeBBox}(M_c)$
    \STATE Generate initial mask: $m_1 \leftarrow S(state, p_1)$

    \STATE Set mask prompt: $p_2 \leftarrow M_c$
    \STATE Generate refined mask: $m_2 \leftarrow S(state, p_2)$

\UNTIL{all images processed}
\end{algorithmic}
\end{algorithm}

\subsubsection{Model Inference}
The inference stage of the FFCL-SAM pipeline is designed to process new, unseen radiographs and produce accurate predictions that identify positive tumor margins. Like the training phase, there are two components to inference using FFCL-SAM: classification and segmentation. In the first step, the trained FFCL model performs a binary classification task on overlapping 64x64 patches extracted from the input radiograph. Each patch is assigned a binary label indicating whether it contains a positive margin or not. The patches, now with this predicted label information, are then mapped back to their original positions in the radiograph, simulating a binary segmentation mask. This mask provides an initial visualization of a potential positive margin but does not lack spatial precision due to the overlapping nature of the patch extraction process.

The second step of inference seeks to address this issue. Based on the rough segmentation mask produced by the patch reconstruction process, two prompts, $p_1$ and $p_2$, are generated and passed to the SAM model, which uses its zero-shot segmentation capabilities to refine the mask. This refinement process generally consists of smoothing the blocky transitions between regions due to the patch-based method and correcting any inconsistencies from the patch-level classifications.

$p_1$ is a bounding box prompt, where the bounding boxes are generated based on the content of the coarse mask. The bounding box generation process can be expressed as:

\begin{equation}
\begin{aligned}
x_{min} = \min\{j : M_c[i,j] = 1 \text{ for some } i\} - \delta_w \\
x_{max} = \max\{j : M_c[i,j] = 1 \text{ for some } i\} + \delta_w \\
y_{min} = \min\{i : M_c[i,j] = 1 \text{ for some } j\} - \delta_h \\
y_{max} = \max\{i : M_c[i,j] = 1 \text{ for some } j\} + \delta_h, \\
\end{aligned}
\end{equation}

\noindent where $M_c$ is the coarse mask, and $\delta_w$ and $\delta_h$ are paddings that are applied to ensure the bounding box accurately captures the whole region of the image containing $M_c$.

$p_2$ is a mask prompt where the entire form of the coarse mask itself is passed to SAM. The purpose of the two-prompt approach employed by FFCL-SAM is to capture the full spatial information of the mask. The goal of $p_1$ is to capture the full area encompassing where the mask is located, and the goal of $p_2$ is to refine SAM'S prediction when $p_1$ was used to better capture the form of the mask. The full process of using SAM 2 to refine the coarse mask is given in Algorithm~\ref{alg:FFCL-SAM}.

\begin{table}[t]
    \centering
    \caption{Distribution of patient radiographic image data and extracted positive/negative margin patches used in our experiments. Only positive cases are used for validation and testing to verify the capability of the model in positive margin detection.}
    \medskip
    \begin{tabular}{lc c c c}
    \toprule
     Patient data & \phantom{a} & Train & Validation & Test \\
     \midrule
     Positive patient && 25  & 2 & 6\\ 
     Negative patient && 13 & – & –\\
     \midrule
     Positive images && 50 & 4 & 12\\
     Negative images && 26 & – & –\\
     \midrule
     Positive patches && 13,207 & 432 & 314\\
     Negative patches && 13,620 & 1,614 & 1266\\
     \bottomrule
    \end{tabular}
    \label{tab:data-dis}
    
\end{table}

\begin{figure}[t]
\centering
\resizebox{\linewidth}{!} {
\begin{tabular}{c c c | c c c}
\includegraphics[width=0.16\linewidth]{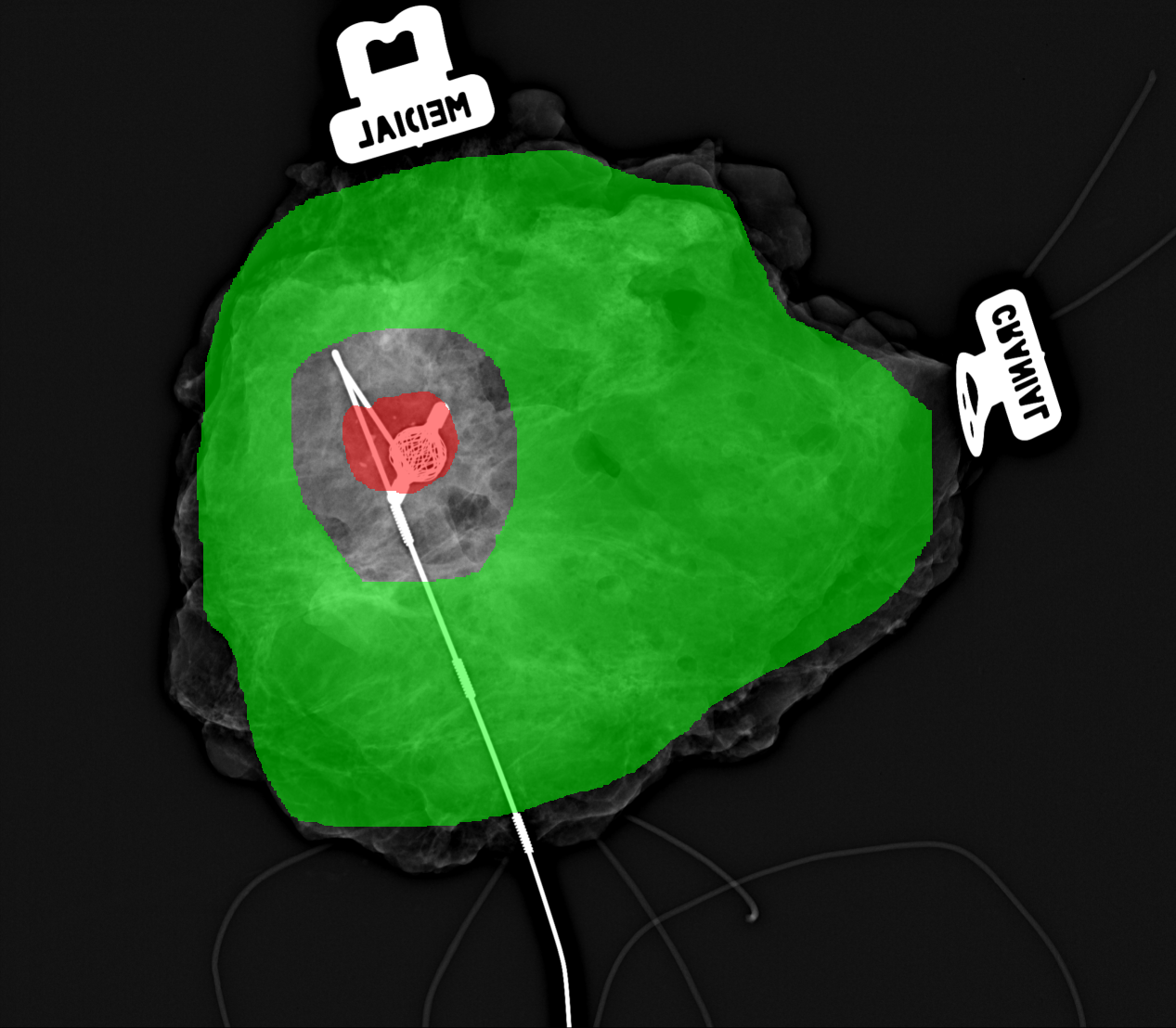} & 
\includegraphics[width=0.16\linewidth]{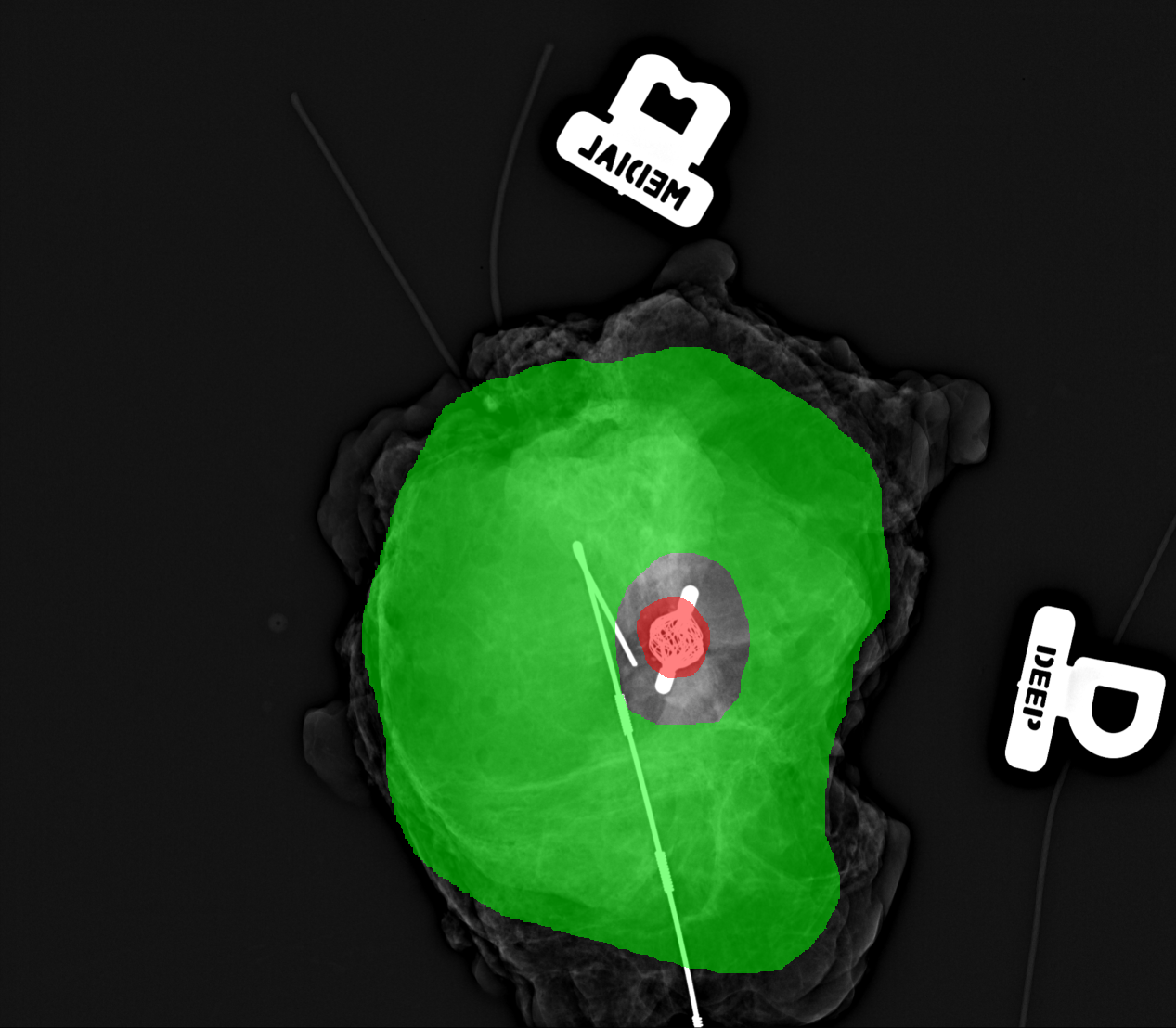} & 
\multicolumn{1}{c}{\footnotesize (a)} &
\includegraphics[width=0.16\linewidth]{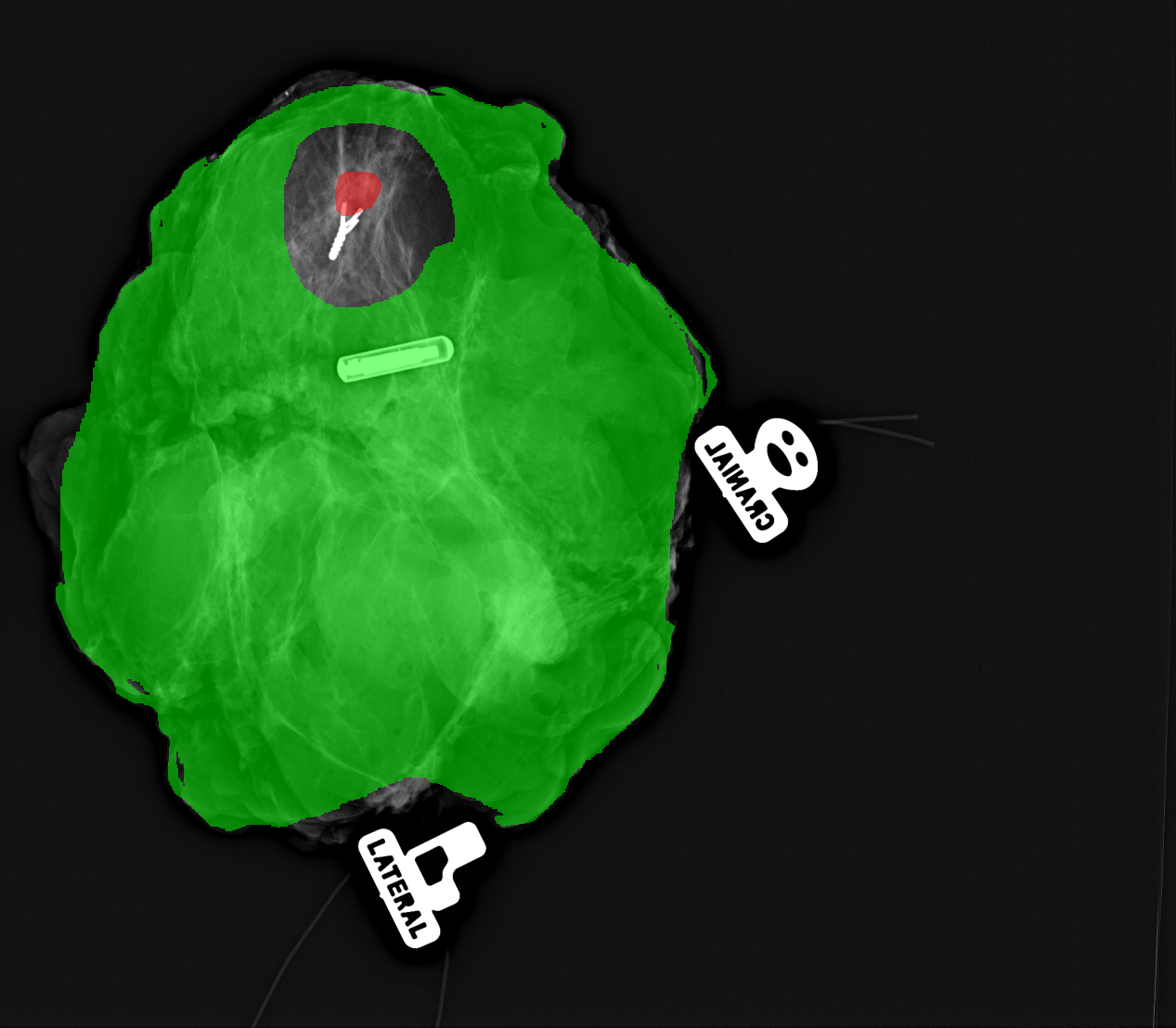} & 
\includegraphics[width=0.16\linewidth]{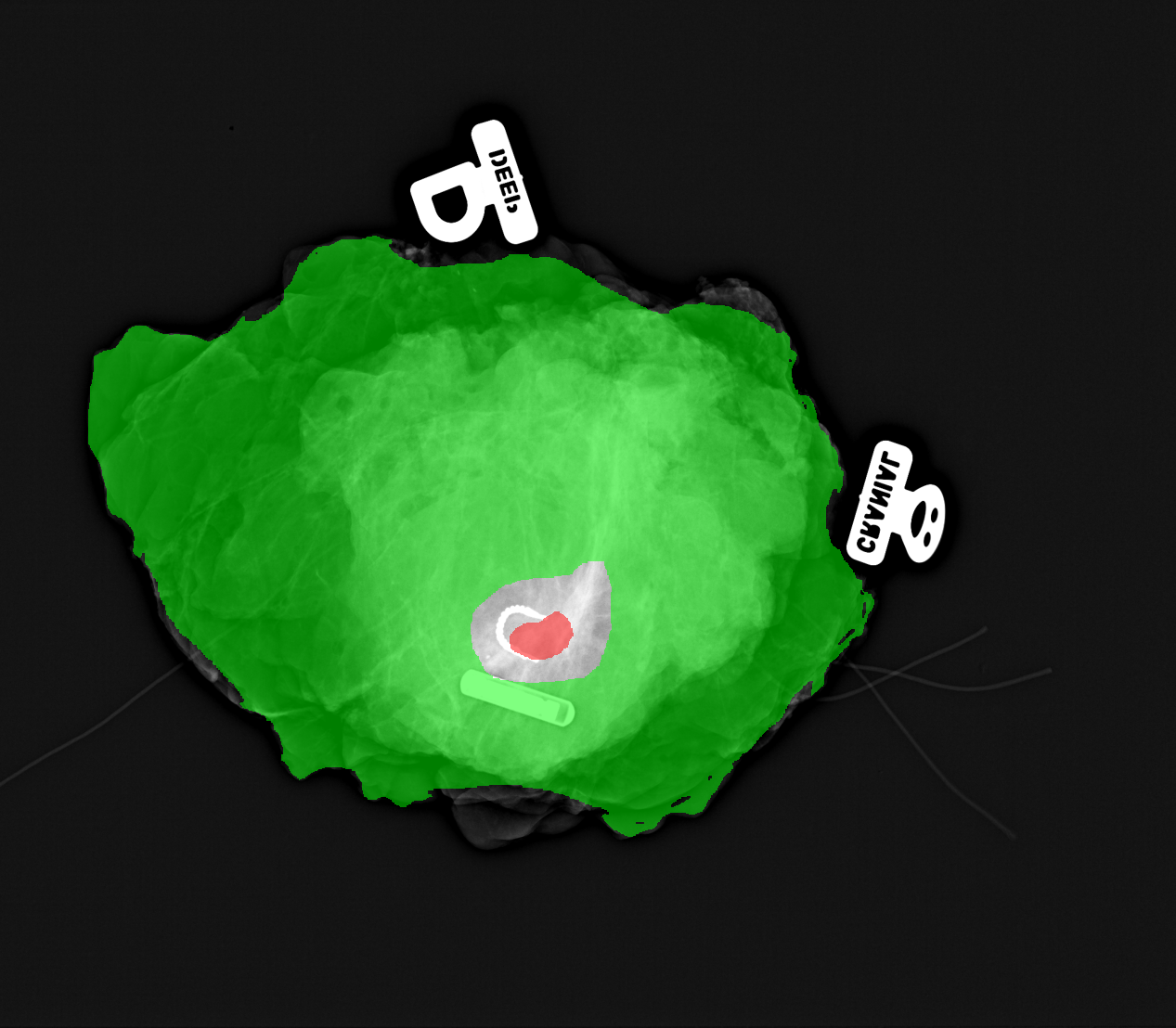} &
\multicolumn{1}{c}{\footnotesize (b)} \\
{\footnotesize Negative Patient AP} & {\footnotesize Negative Patient CC} & & {\footnotesize Negative Patient CC} & {\footnotesize Negative Patient MLO} & \\
{\footnotesize Margin Absent} & {\footnotesize Margin Absent} & & {\footnotesize Margin Absent} & {\footnotesize Margin Absent}\smallskip & \\
\includegraphics[width=0.16\linewidth]{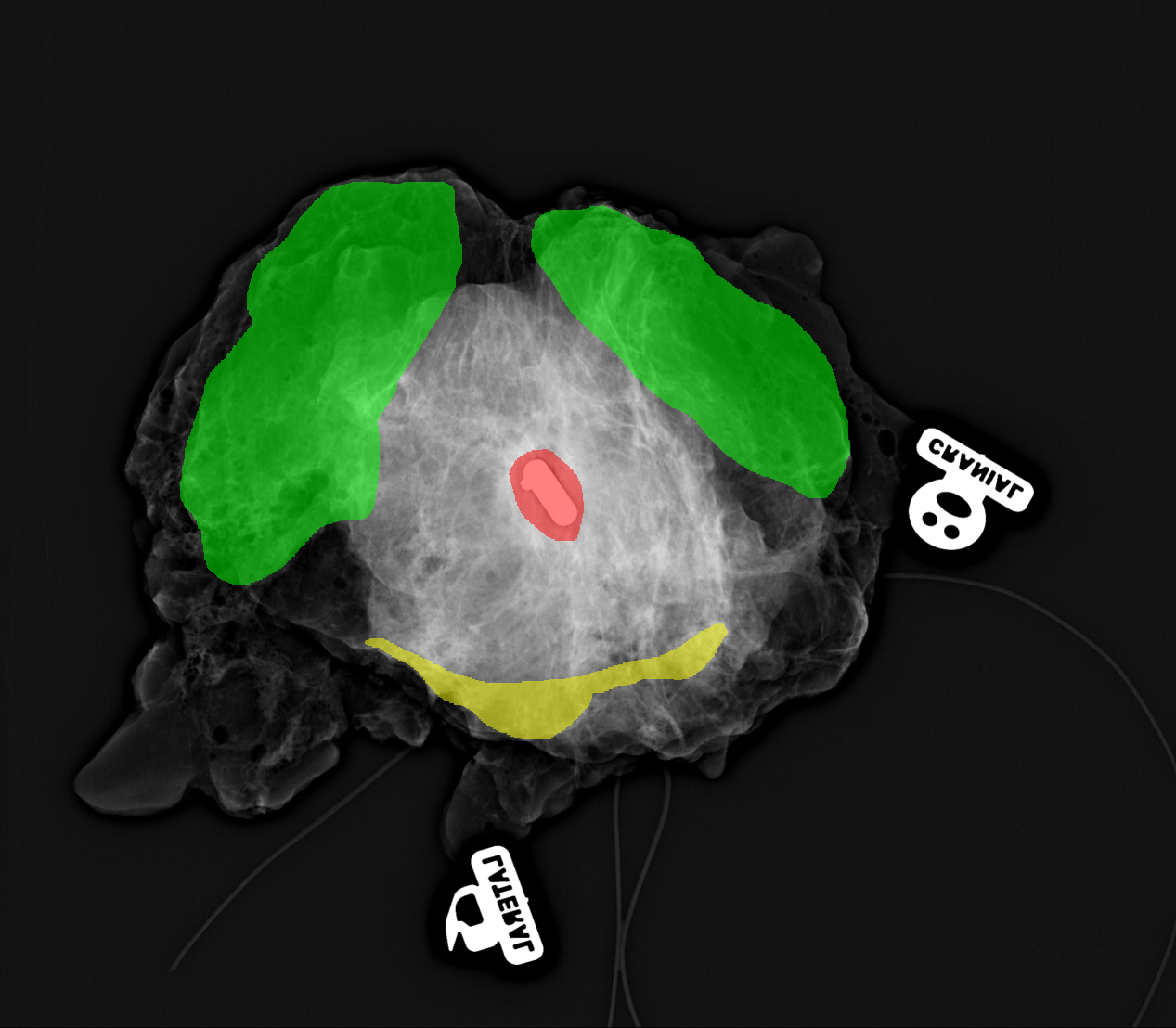} & 
\includegraphics[width=0.16\linewidth]{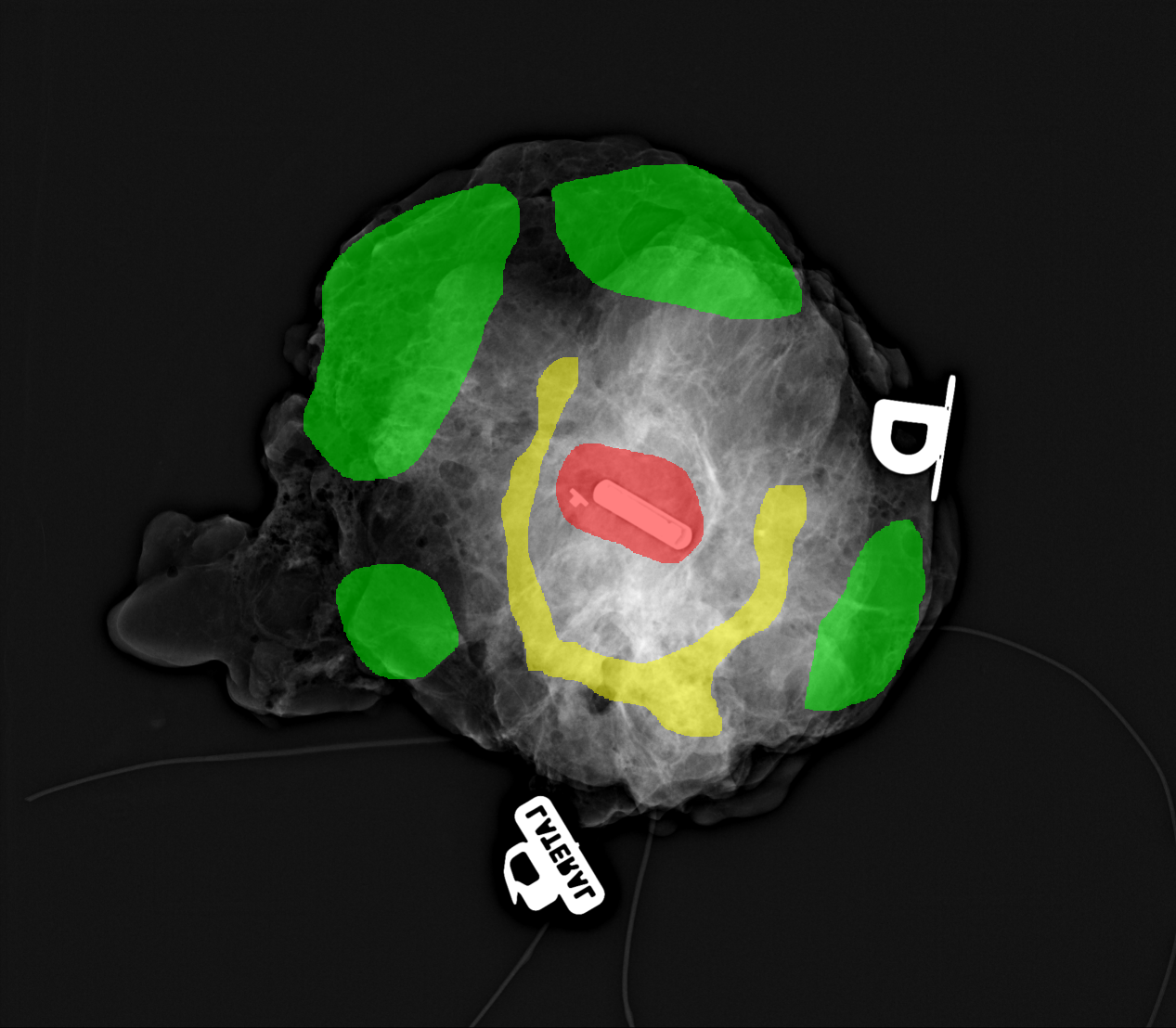} &
\multicolumn{1}{c}{\footnotesize (c)} &
\includegraphics[width=0.16\linewidth]{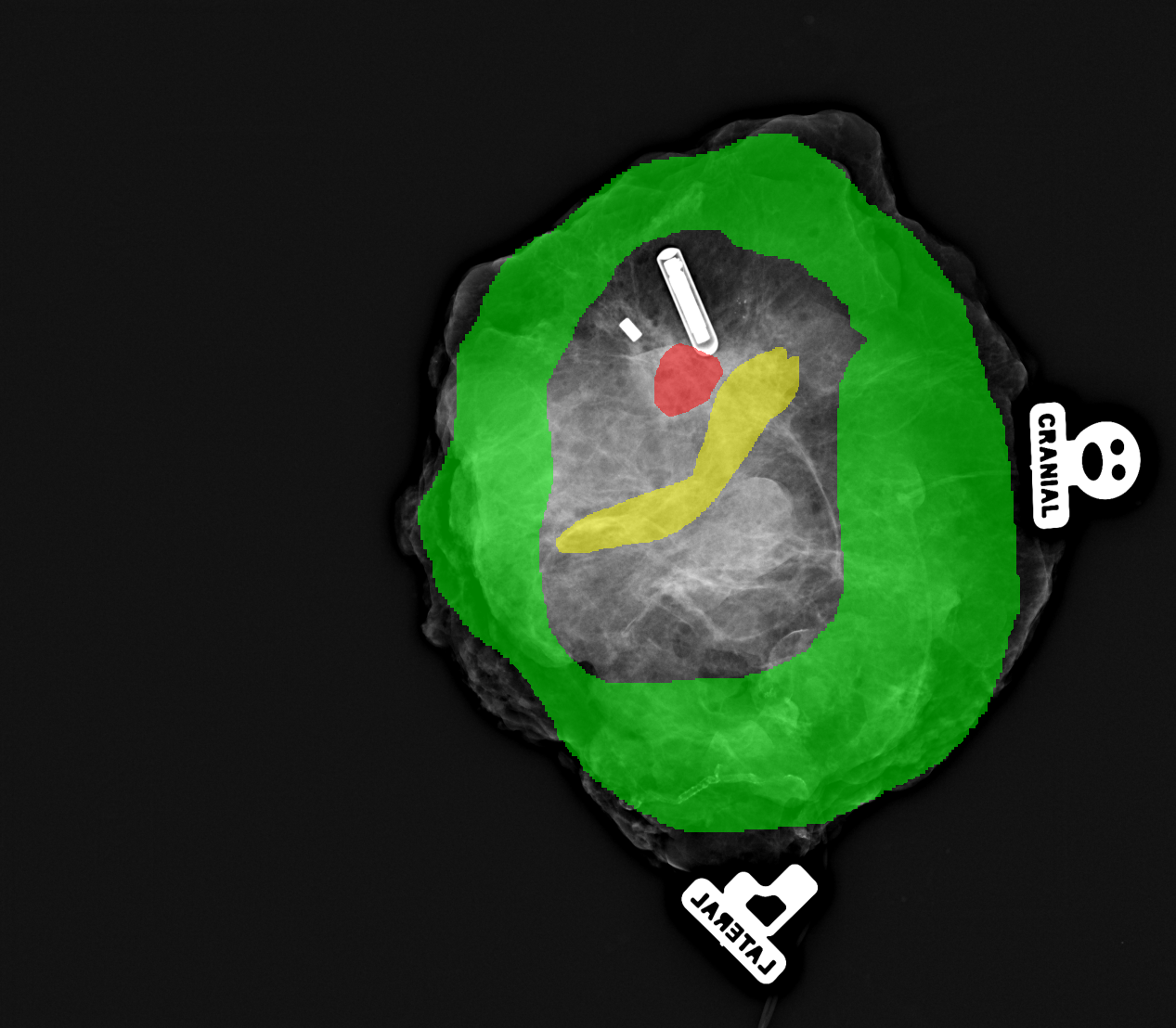} & 
\includegraphics[width=0.16\linewidth]{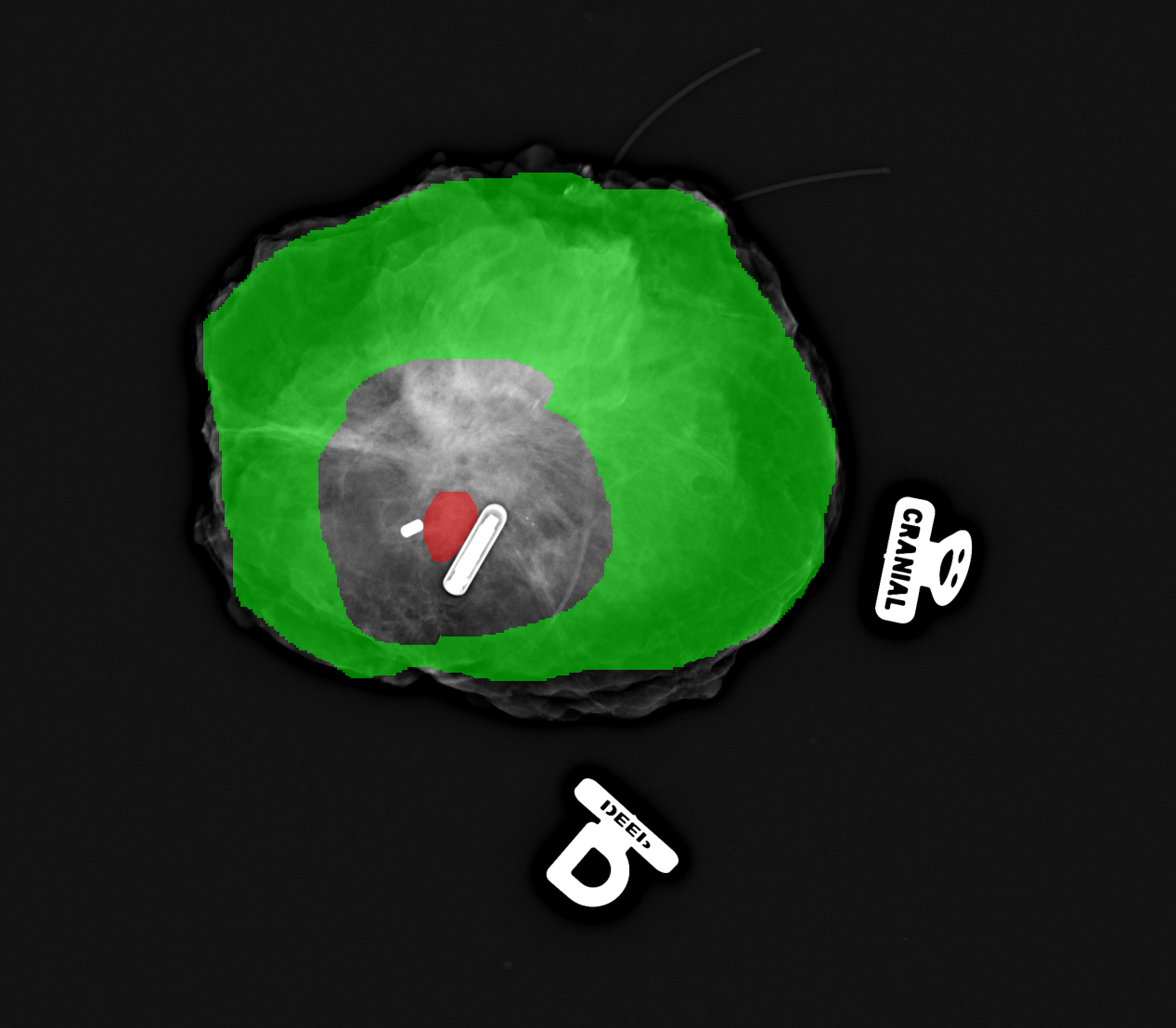} &
\multicolumn{1}{c}{\footnotesize (d)} \\
{\footnotesize Positive Patient AP} & {\footnotesize Positive Patient CC} & & {\footnotesize Positive Patient AP} & {\footnotesize Positive Patient MLO} & \\
{\footnotesize Margin Present} & {\footnotesize Margin Present} & & {\footnotesize Margin Present} & {\footnotesize Margin Absent} &
\end{tabular}
}
\caption{Anterior-posterior (AP), craniocaudal (CC), and mediolateral (MLO) image samples from our dataset with their corresponding segmentation masks. Each patient has two image samples of different views associated with them. (a) and (b) are taken from the negative patient data set, while (c) and (d) are taken from the positive set. Observe that in the negative set, no sample contains a positive margin, while in the positive set, some views may not contain a positive margin, but there is no case where both image views for a patient would be without a positive margin. Color code: Green - non-malignant tissue, yellow - positive margin, red - malignancy.}
\label{fig:masks}
\end{figure}

\section{Experimental Evaluation}
\label{sec:exp&results}

\subsection{Data and Annotation}

In this HIPAA (Health Insurance Portability and Accountability Act)-compliant retrospective study, data was collected from intraoperative specimen radiographs from a pool of 215 localized lumpectomies with approval from the Institutional Review Board (IRB). The lumpectomies were performed between 2019 and 2022 at the University of Kentucky and Markey Cancer Center Comprehensive Breast Care Center. Cases without anatomic orientation, labeled intraoperative radiographs, or pathology-confirmed malignancy within the specimen were excluded. Diagnostic mammograms and post-biopsy images were reviewed to localize the tumor within each specimen and 3-D orientation of the 2D radiographs. Pathology reports from the lumpectomy were reviewed to assess the margin status for the main specimen, which was imaged intraoperatively. Lumpectomies with multiple submitted specimens were reviewed by pathologists to confirm the margin status of the main imaged specimen. 

We use the open-source ITK-SNAP tool to annotate the lumpectomy radiograph images, denoting regions of known malignancy, non-malignant tissue, and the area of pathology-confirmed positive margin. 
The radiograph images were de-identified and converted to NIfTI file format for image annotation within ITK-SNAP~\cite{yushkevich2016itk}. The specimen radiograph was annotated with 3 distinct region labels on the annotation mask (Fig.~\ref{fig:masks}). Region Label 1 represents the area of the lumpectomy specimen that contains nonmalignant tissue, named “Negative”. Region Label 2 represents the area of the specimen that contains the observable tumor within the specimen, named “Positive”. Region Label 3 represents the margin along which the pathology assessment was deemed positive, named “Positive Margin”. Margin status was deemed positive or negative following the current guidelines of the Society of Surgical Oncology, American Society of Clinical Oncology, and American Society of Radiation Oncology~\cite{moran2014society}. Lumpectomy specimens containing invasive carcinoma with or without Ductal carcinoma in situ (DCIS) were considered negative with no ink on tumors, and specimens containing pure DCIS were considered negative for margins greater than 2mm.

In our experiments, we use 92 radiographs from 46 patients and split them into train, validation, and test sets as reported in Table~\ref{tab:data-dis}. Considering the small dataset, we enlarge the three sets by extracting a large number of patches from the positive and negative regions. Please note that, for robust analysis, patches are extracted from the positive margin regions only for the validation and test sets. This approach while adds a challenge for the model, but at the same time, it encourages the models to learn better generalized feature representations. 

\subsection{Implementation Details}
\textbf{Training:}
We explore with two CNN (ResNet-18 \cite{he2016deep} and ConvNeXt \cite{liu2022convnet}) and a transformer (ViT \cite{dosovitskiy2020image}) architectures in our FFCL experiments.

\begin{figure}
    \centering
    \includegraphics[width=0.7\linewidth, trim={0cm 0cm 0cm 0cm},clip]{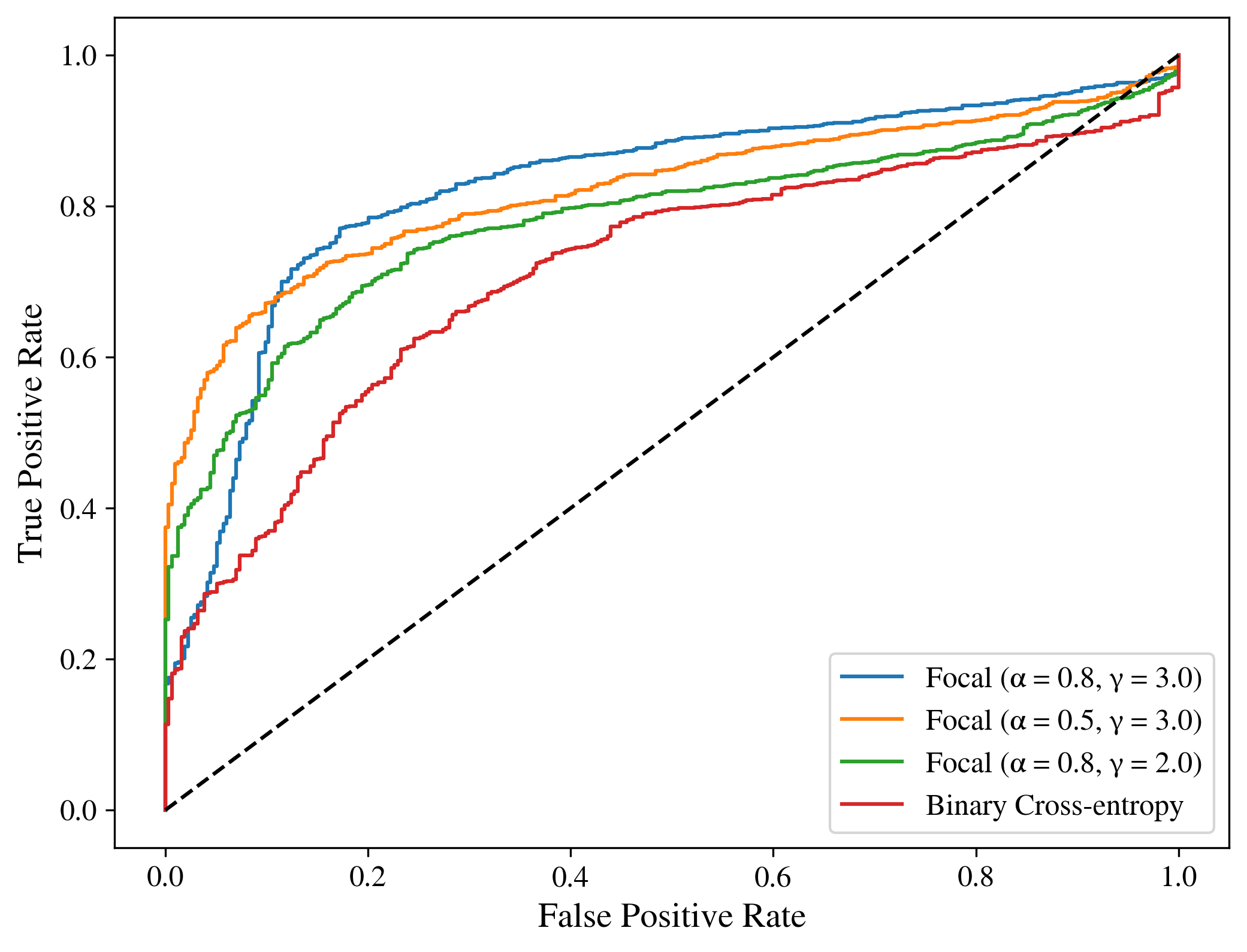}
    \caption{Visualization of Receiver operating curve (ROC) curves comparing the binary cross-entropy and Focal cross-entropy losses (w/ ResNet-18) at different $\alpha$ and $\gamma$ parameters.}
    \label{fig:roc}
    \vspace{4pt}
\end{figure}

We leverage the CNN or transformer backbone as the feature extractor by removing the last fully connected (FC) layer, which outputs a 512-dimensional feature vector. We add an FC layer with 1 neuron followed by sigmoid, to make the binary class (positive/negative) prediction. We adopt the Focal cross-entropy loss to focus on a selective set of challenging examples, preventing the abundance of straightforward negatives from overpowering the detector during its training phase. For each experiment, an $\alpha$ of 0.8 and a $\gamma$ of 3.0 was used for the focal loss. We came to these values following a series of experiments to determine the best-performing values, the results of which are shown in Fig.~\ref{fig:roc}). 64$\times$64 patches are fed to FFCL as input. We train the model with a batch size of 10. A balanced set of positive and negative patches is loaded per batch during training to avoid a biased prediction. We train our model with 50 epochs for all the experiments. We use the Cosine Annealing learning rate scheduler and Adam optimizer with an initial learning rate set to 0.0001. The backbone architectures are taken with randomly initalized weights to train and evaluate the positive margin detection task. After FFCL-based contrastive pre-training, the downstream classification models are trained for 50 epochs with an early stopping of patience=5, and the model is saved based on the best validation loss. 

Once the initial margin masks are reconstructed by assigning positive/negative class labels to the patches, we use SAM to refine and obtain the final BCS margins. In this work, we explore the use of SAM and its successor, SAM 2. For our SAM experiments, we use the ViT-B backbone, which contains 91M parameters. For SAM 2, we use the Hiera-Small \cite{ryali2023hiera} backbone, which contains 46M parameters. 

\textbf{Machine Configuration:} FFCL-SAM is implemented in Python with PyTorch and run on an Intel(R) Xeon(R) W7-2475X processor (2600MHz) with 128GB RAM and dual NVIDIA307 A4000X2 GPUs (32GB).

\textbf{Baseline and Compared Methods:} For the classification task, we evaluate three variants of the proposed FFCL-SAM, which uses ResNet-18, ConvNeXt-tiny, or ViT-b-16 as the backbone. For margin segmentation, we explore both versions of SAM (SAM and SAM2) in zero-shot settings and compare them against the baseline FFCL-reconstructed margin masks.

\subsection{Evaluation Measures}
For comprehensive evaluation of classification and segmentation performance of the proposed FFCL-SAM, we calculate and report several performance metrics.
In evaluating classification, we first obtain a confusion matrix and extract true positive (TP), true negative (TN), false positive (FP), and false negative (FN) values for all the models. We then calculate accuracy, precision, recall, F1, and area under the receiver operating characteristics (ROC) curve as follows.

\begin{itemize}
\item \textbf{Accuracy:} The proportion of correct predictions out of all predictions. It ranges between 0 and 1 (higher is better).
\begin{equation}
Accuracy = \frac{TP+TN}{TP+TN+FP+FN}
\end{equation}
\item \textbf{Precision:} The number of positive predicted instances that are actually positive. It ranges from 0 to 1, where a score of 1 indicates perfect precision and 0 represents no precision at all. 
\begin{equation}
Precision = \frac{TP}{TP + FP}.
\end{equation}
\item \textbf{Recall:} The number of actual positive instances that are correctly predicted. Recall score ranges from 0 to 1 (higher is better). 
\begin{equation}
Recall = \frac{TP}{TP + FN}.
\end{equation}
\item \textbf{F1:} The harmonic mean of precision and recall. F1 scores range from 0 to 1, where 0 is the worst possible score and 1 is the best.
\begin{equation}
F1 = 2 \times \frac{Precision \times Recall}.{Precision + Recall}
\end{equation}
\item \textbf{AUC:} Overall success of a classifier
\begin{equation}
AUC = \int_0^1 Pr[TP] vdv,
\end{equation}
where $Pr[TP]$ (true positive rate) is a function of $v=Pr[FP]$ (false positive rate). AUC ranges from 0.5 to 1. AUC values higher than 0.8 are generally considered clinically useful~\cite{ccorbaciouglu2023receiver}.
\end{itemize}

For evaluating the segmentation performance, we calculate the standard Dice similarity coefficient (DSC), Hausdorff distance (HD), and accuracy scores across. For $A$ denoting the set of pixels in the predicted mask and $B$ is the set of pixels in the ground truth mask, DSC and HD can be calculated as follows. 

\begin{itemize}
\item \textbf{Dice Similarity Coefficient (DSC):} Measures the overlap between two sets, commonly used in medical image segmentation to compare the predicted segmentation mask with the ground truth mask. DSC ranges between 0 and 1, where 0 means no overlap and 1 denotes a perfect overlap between the ground truth and prediction.
\begin{equation}
DSC = \frac{2 \times |A \cap B|}{|A| + |B|}.
\end{equation}
\item \textbf{Hausdorff Distance (HD):} Measures the worst-case difference between two sets of points, capturing how far apart the predicted segmentation boundary is from the ground truth. The lower the HD is, the better.
\begin{equation}
H(A, B) = \max \left( \sup_{a \in A} \inf_{b \in B} d(a, b), \sup_{b \in B} \inf_{a \in A} d(a, b) \right),
\end{equation}
\end{itemize}

\noindent where $|A \max B|$ is the number of pixels where the prediction and ground truth overlap. $d(a, b)$ is the Euclidean distance between points $a$ and $b$.

\begin{table}[t]
    \centering
    \caption{Quantitative evaluation of the FFCL-based pretraining of different deep learning architectures in detecting the lumpectomy margins. Average$\pm$ standard deviation for five evaluation metrics (Accuracy, F1, Precision, Recall, and AUC) have been reported after 5 runs of each of the models. Best scores are \textbf{bolded} and Second best are \underbar{underlined}.}
    \medskip
    \resizebox{\linewidth}{!}{
    \setlength{\tabcolsep}{4pt}
    \begin{tabular}{lc c c c c c}
    \toprule
     Model & \phantom{a} & Accuracy & F1 & Precision & Recall & AUC\\
     \midrule
    ConvNeXt w/ FFCL && 0.7510 ± 0.0029 & 0.6396 ± 0.0100 & 0.6542 ± 0.0186 & \underline{0.6322 ± 0.0072} & 0.7751 ± 0.0002\\
     ViT w/ FFCL && \textbf{0.7816 ± 0.0316} & \underline{0.6898 ± 0.0373} & \textbf{0.7882 ± 0.0582} & 0.6006 ± 0.0387 & \underline{0.8067 ± 0.0530}\\
     ResNet-18 w/ FFCL && \underline{0.7815 ± 0.0174} & \textbf{0.6914 ± 0.0238} & \underline{0.7478 ± 0.0656} & \textbf{0.6900 ± 0.0142} & \textbf{0.8455 ± 0.0152}\\
     \bottomrule
    \end{tabular}
    }
    \label{tab:pretrainedresults}
\end{table}

\subsection{Results and Discussion}
\paragraph{Margin detection via classification (FFCL):}
FFCL was previously shown to be more effective than other existing contrastive learning strategies~\cite{ahamed2023ffcl, ahamed2024automatic}.
Table~\ref{tab:pretrainedresults} reports comparisons between CNN and transformer-based models pre-trained using FFCL. For fair comparisons, all three architectures (ResNet-18, ViT, and ConvNeXt) were randomly initialized, which is also consistent with the optimal setting in our prior findings~\cite{ahamed2023ffcl}. FFCL demonstrates the effectiveness in detecting lumpectomy margins from SRs, outperforming its non-pretraining counterparts with the same architecture (e.g., ResNet-18 w/ FFCL over ResNet-18 only)~\cite{ahamed2024automatic}.  
Among the three architectural choices, ResNet18 w/ FFCL has the best overall performance, achieving the best results across three of the five metrics and the second-best in the other two. According to the literature, an AUC score higher than 0.8 strongly suggests the potential clinical applicability of the model~\cite{ccorbaciouglu2023receiver}. Moreover, ResNet-18 w/ FFCL outperforms existing RadImageNet pre-trained InceptionV3~\cite{chen2023analysis} in terms of AUC.
ViT w/ FFCL performs comparably, in cases surpassing in terms of accuracy by 0.01\% and precision value by 3.64\%. However, compared to the ResNet-18 model, ViT takes significantly longer to train (32m and 16.33m for local and global contrastive pretraining, 8.5 hours for FFCL pretraining compared to ResNet18's 5m, 16.33m, and 50m), has a longer inference time (771ms vs. 97ms) and requires much more memory (23MB vs 391MB) and computational power for training/inference. The model size of ViT (86M) compared to ResNet-18 (11M) and ConvNeXt (28M) is another limiting factor. This also indicates that using ViT could be risky in certain scenarios, as the more parameters a model has, the higher the likelihood that it can overfit to the training data, especially on such small datasets as the ones commonly used for margin assessment tasks.
The absence of actual positive margin patches in training further showcases the capability of FFCL in intrinsic margin features. Considering the superiority of ResNet-18 in terms of efficiency and effectiveness in detecting positive margins, the subsequent experiments are performed using ResNet-18. 

\paragraph{Ablation Experiments:}
In order to determine the optimal settings for the Focal loss in classification, we performed an ablation experiment with different combinations of the $\alpha$ and $\gamma$ parameters. Fig.~\ref{fig:roc} compares the ROC curves for the three settings of the Focal loss as well as a binary cross-entropy loss. Based on the ROC curve, it is evident that the Focal loss with an $\alpha$ of 0.8 and a $\gamma$ of 3.0 achieves the best performance among all four. We, therefore, select this as the optimal setting and continue the following FFCL-based margin detection experiments.

\begin{figure}[t]
\centering
\resizebox{0.85\linewidth}{!} {
\begin{tabular}{c c c c c}
{Input SR} & {Ground Truth} & {Coarse Mask} & {SAM-refined mask} & {SAM2-refined Mask} \\
\includegraphics[width=0.24\linewidth]{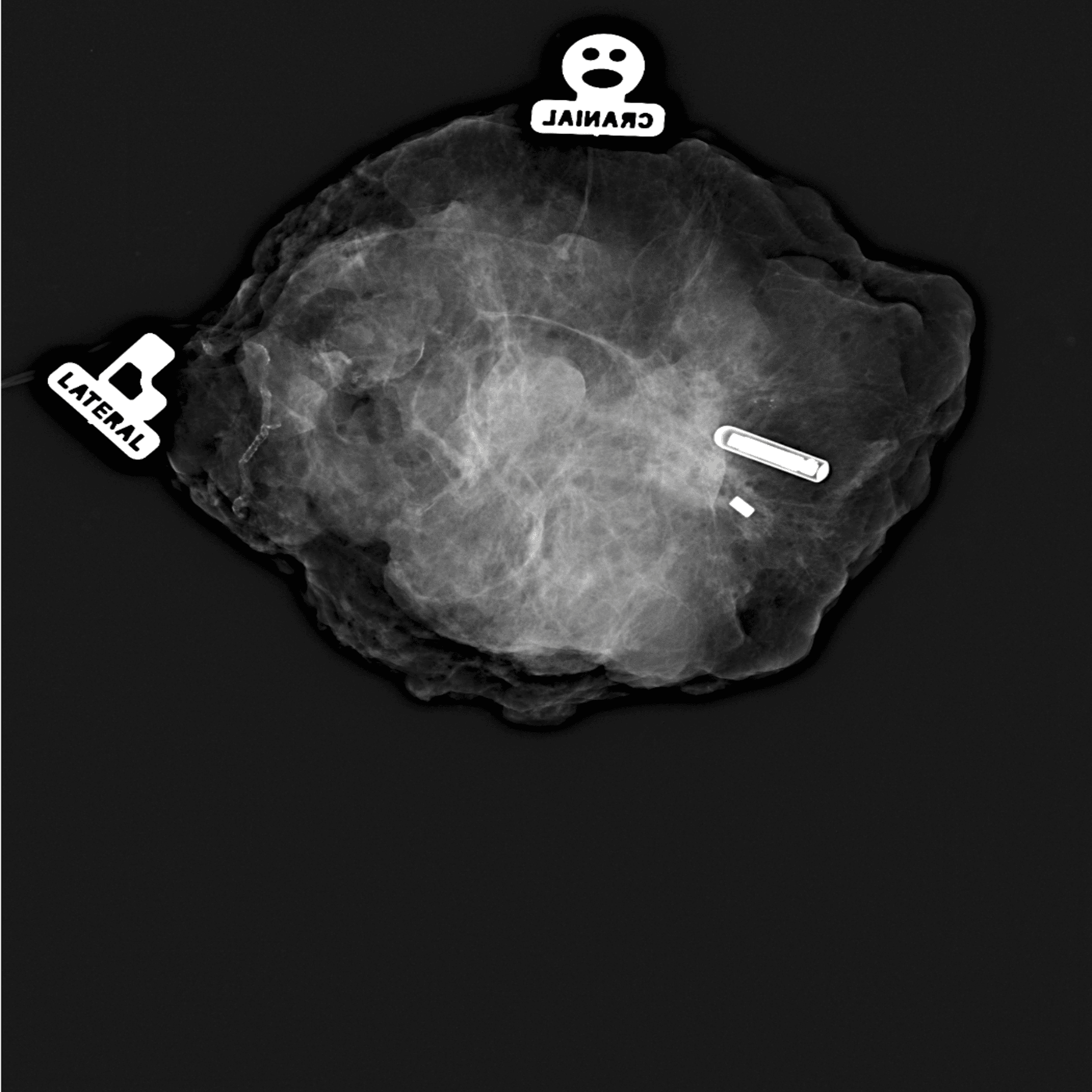} & 
\includegraphics[width=0.24\linewidth]{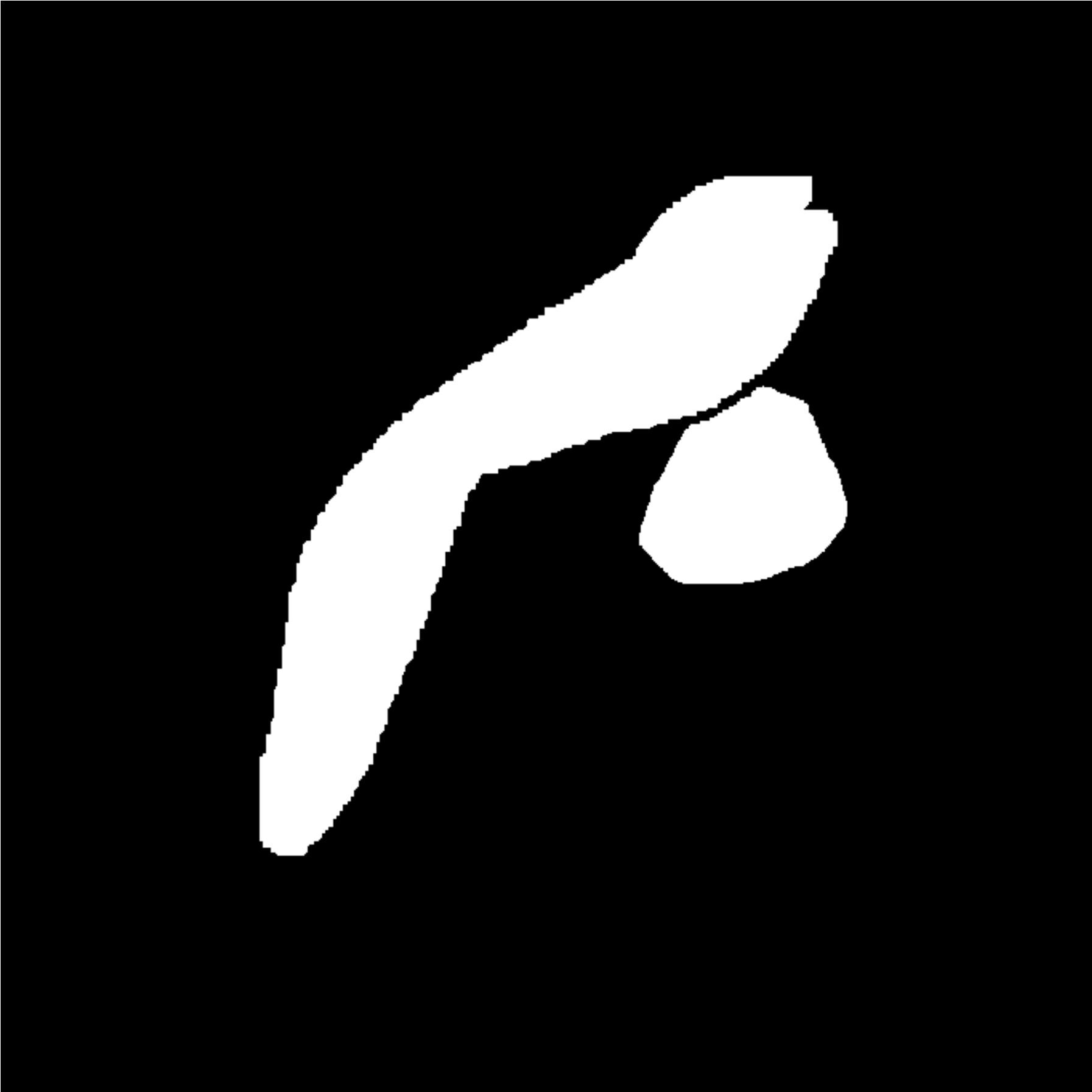} & 
\includegraphics[width=0.24\linewidth]{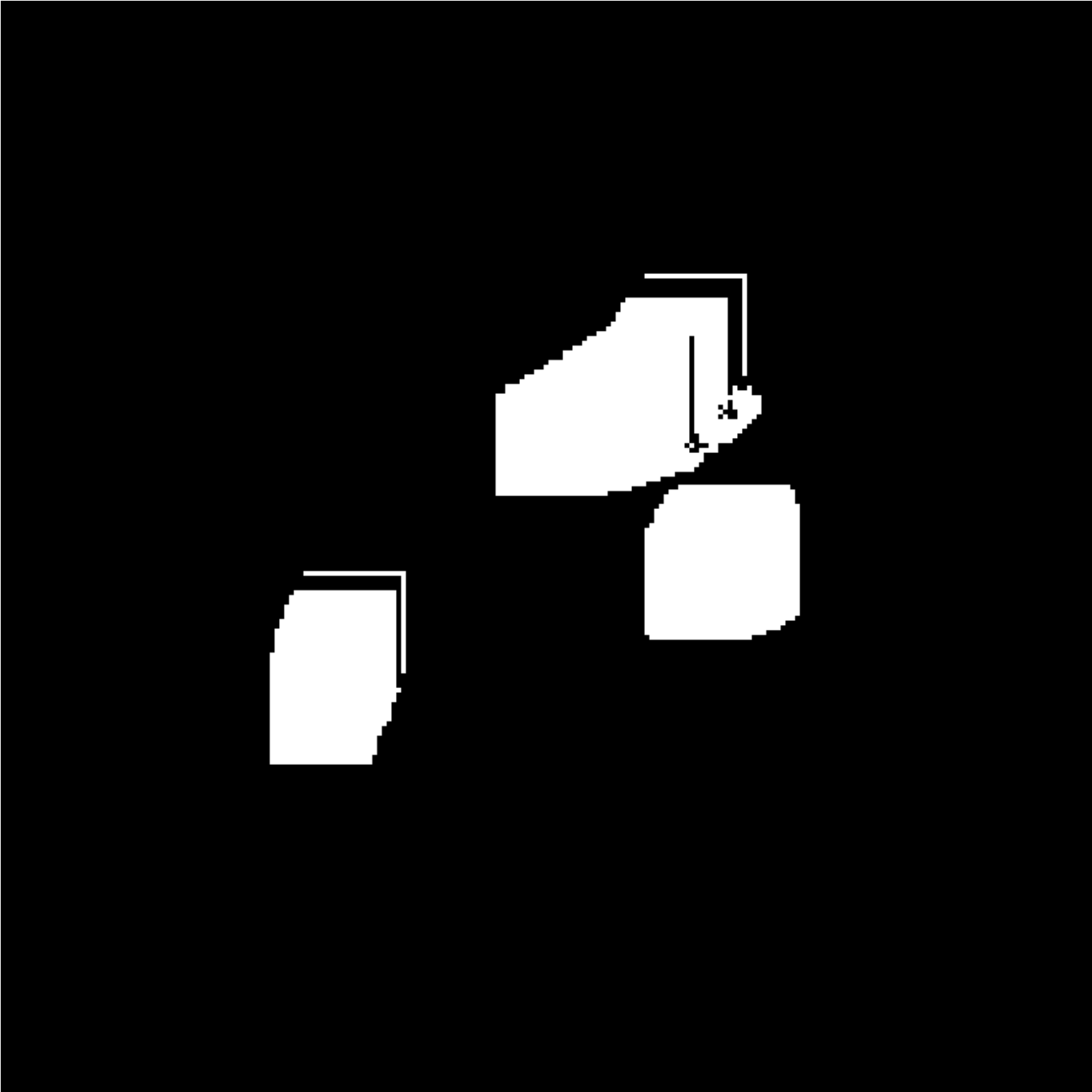} &
\includegraphics[width=0.24\linewidth]{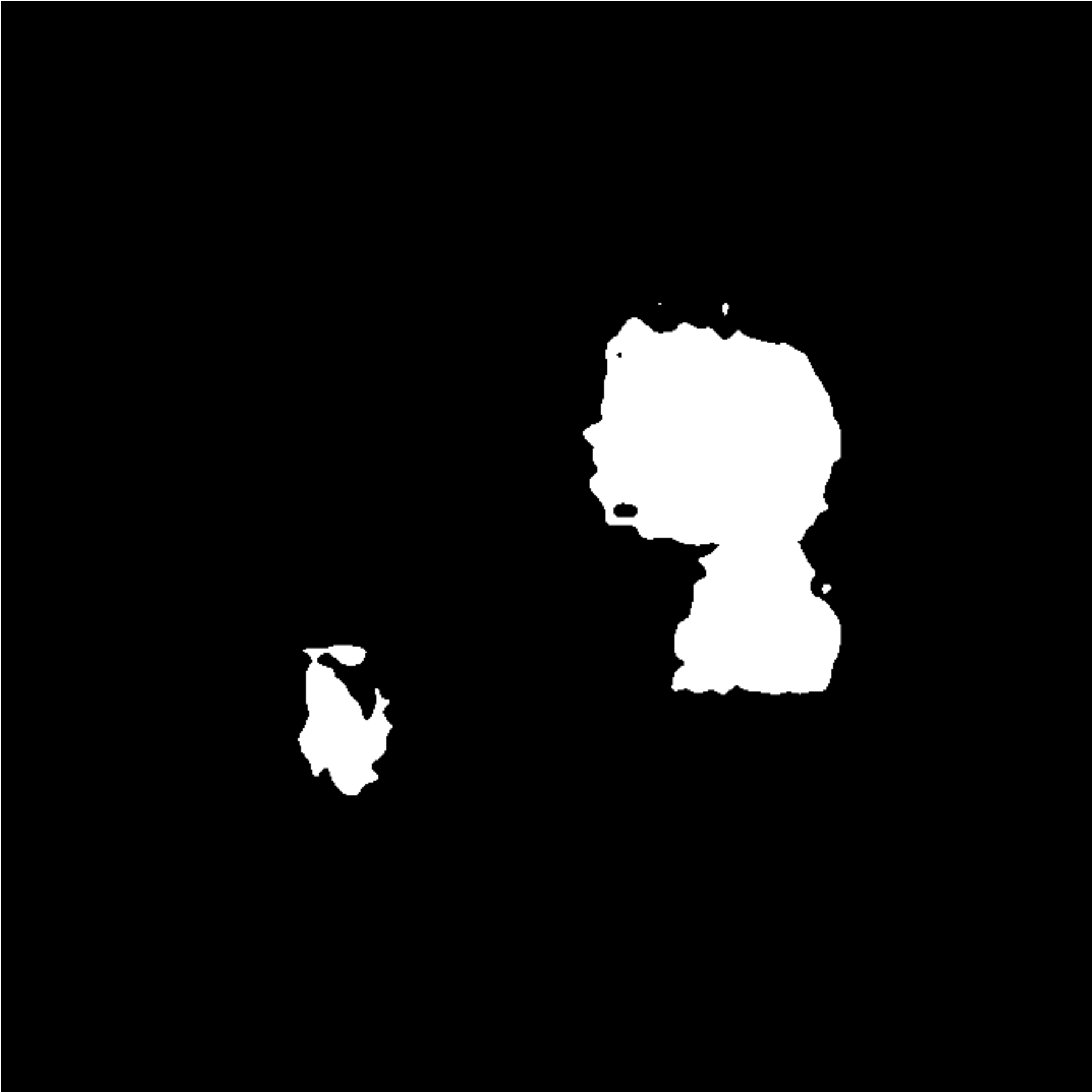} &
\includegraphics[width=0.24\linewidth]{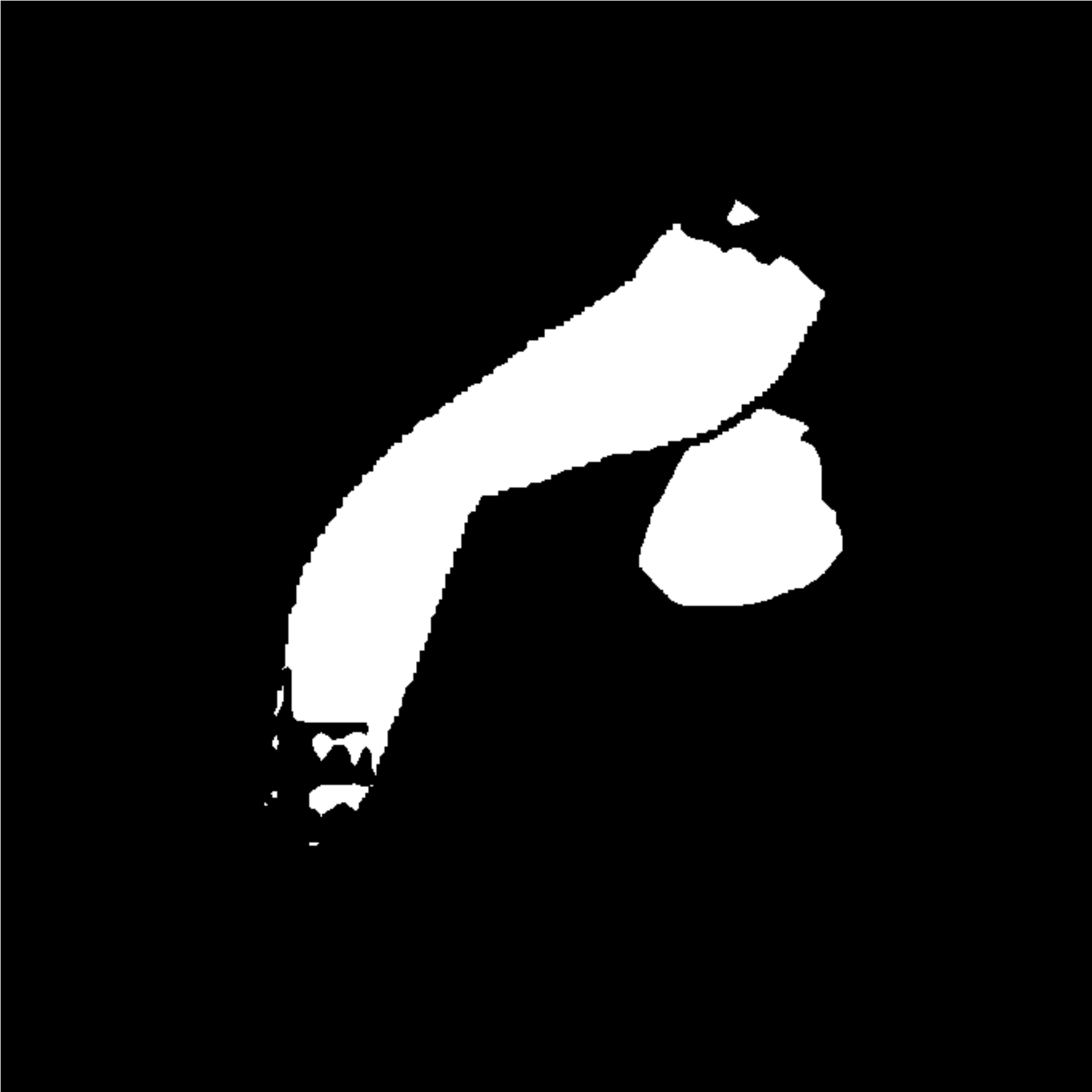} \\

\includegraphics[width=0.24\linewidth]{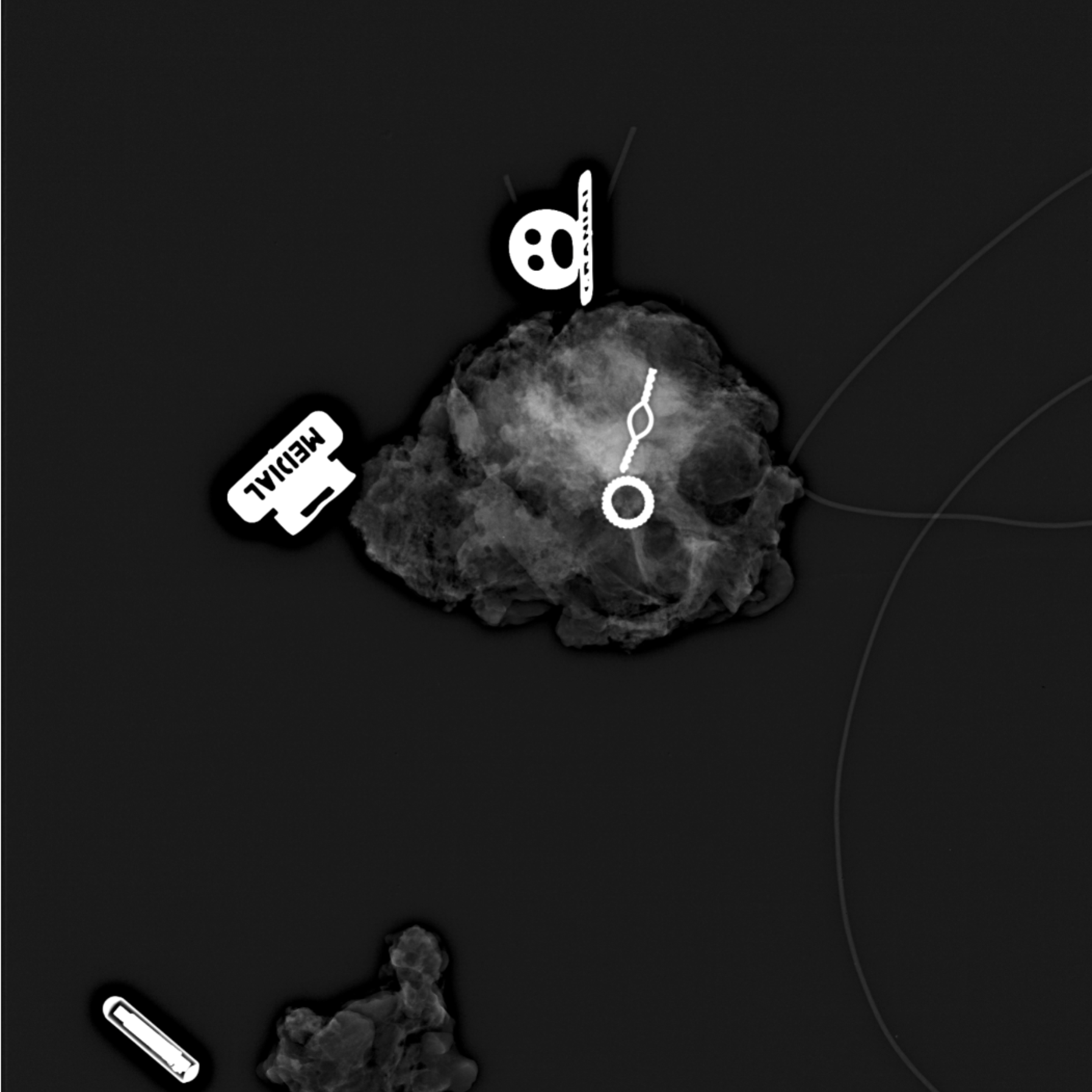} & 
\includegraphics[width=0.24\linewidth]{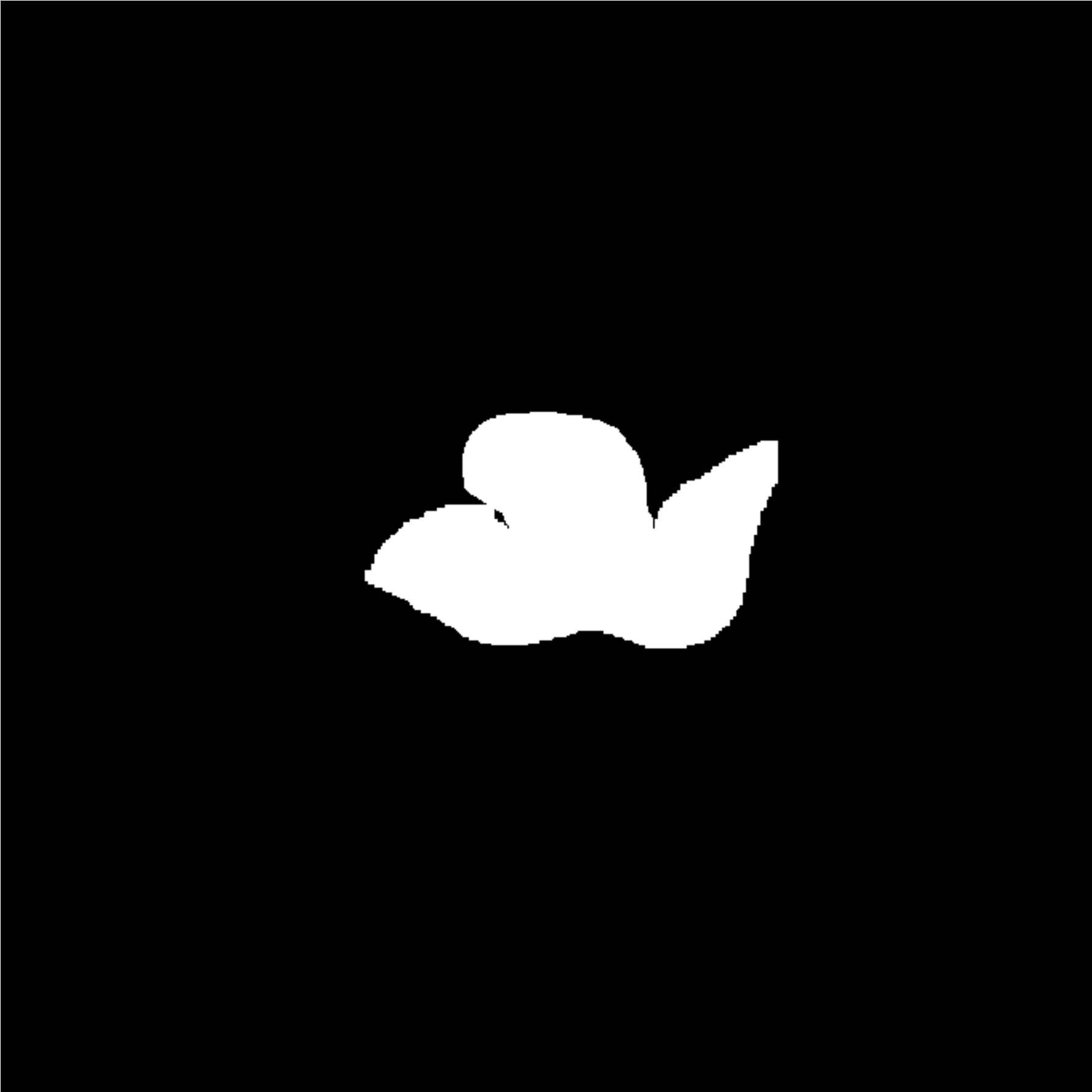} & 
\includegraphics[width=0.24\linewidth]{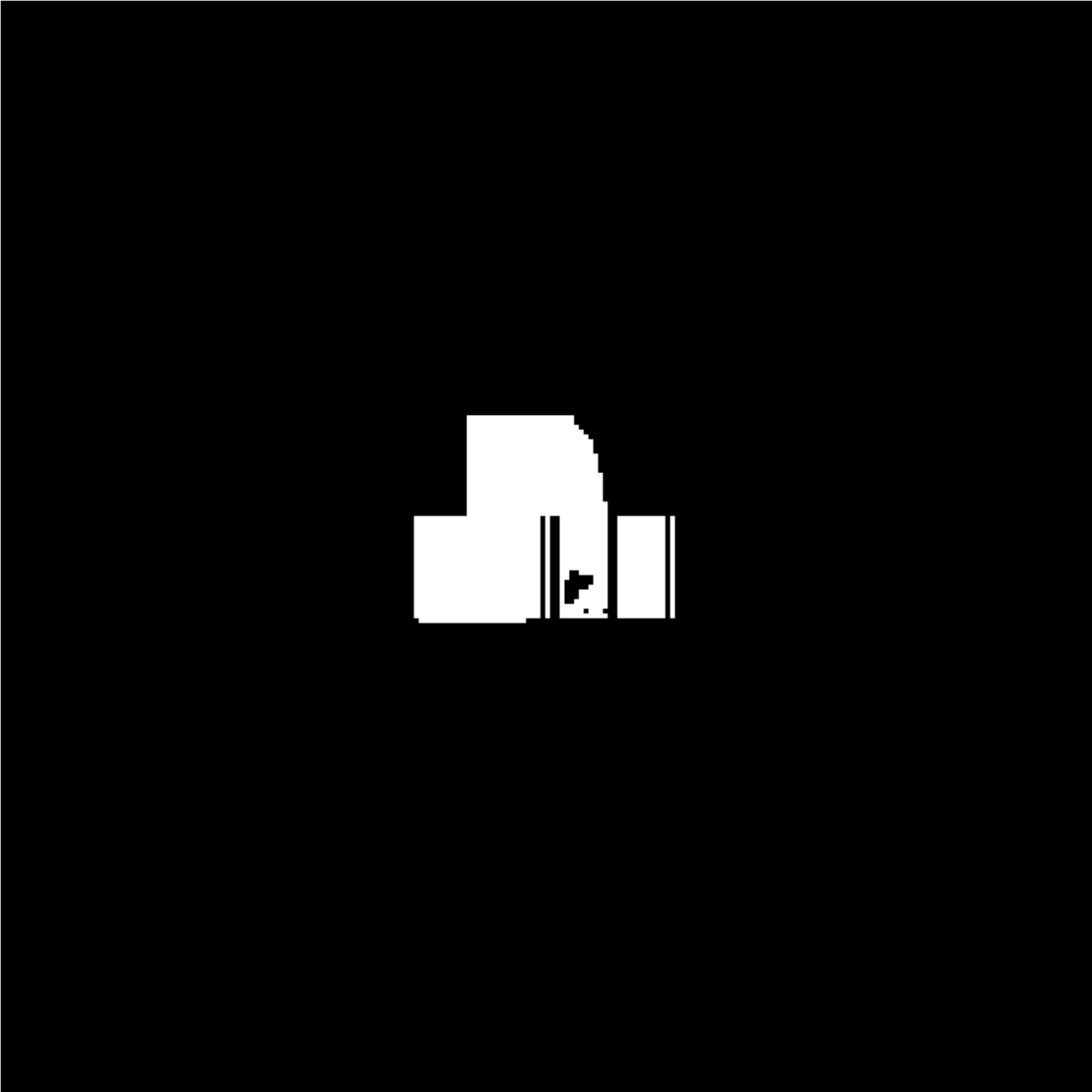} &
\includegraphics[width=0.24\linewidth]{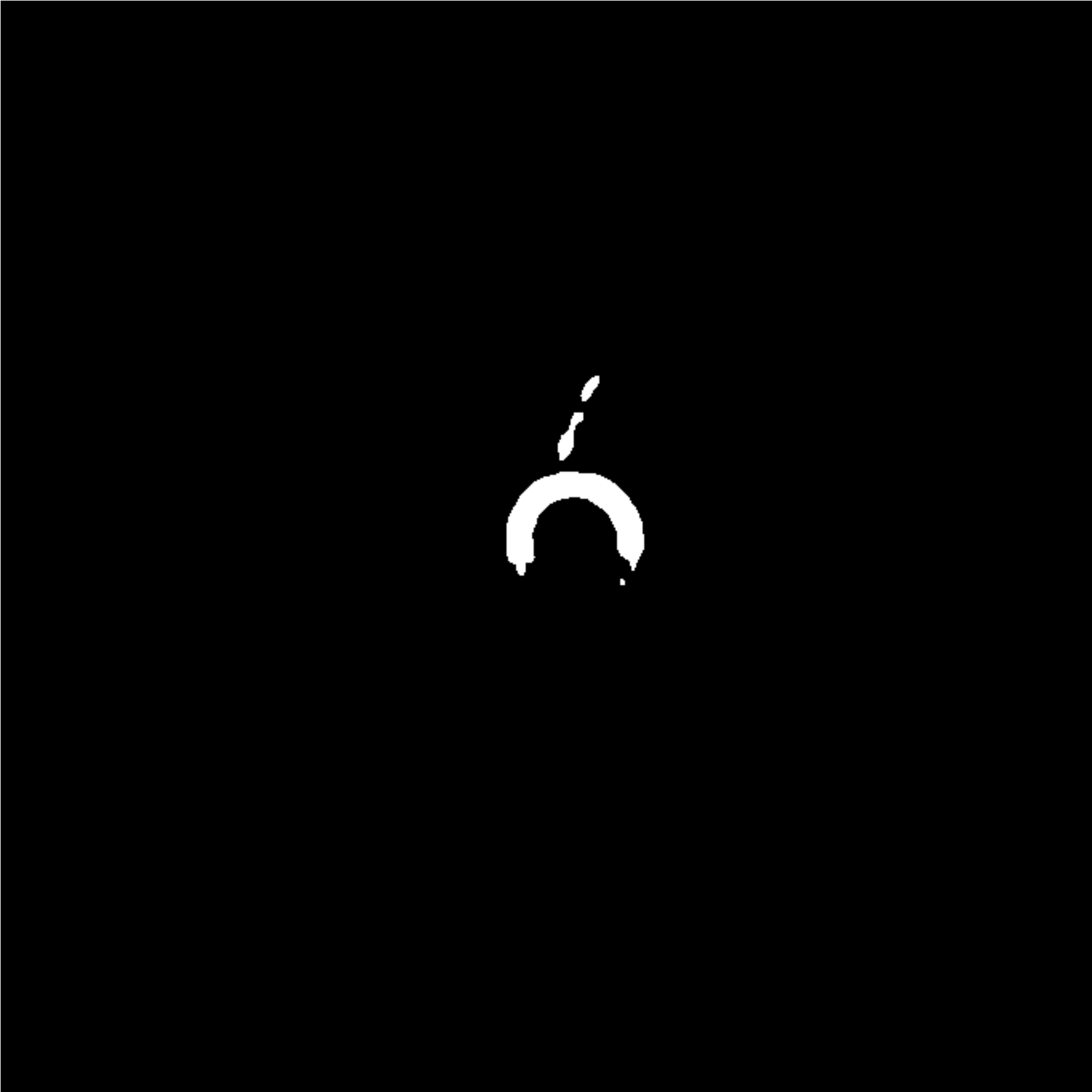} &
\includegraphics[width=0.24\linewidth]{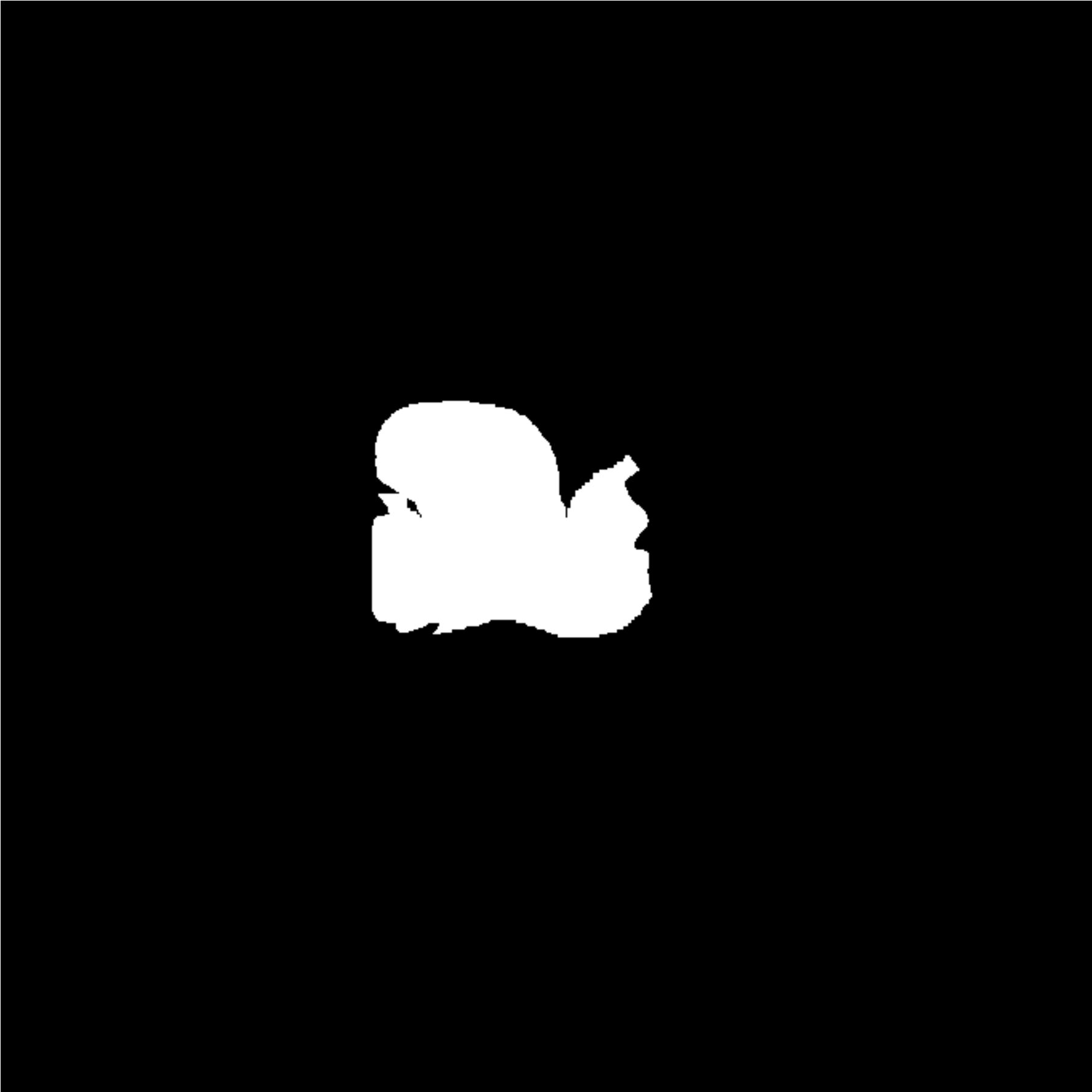} \\

\includegraphics[width=0.24\linewidth]{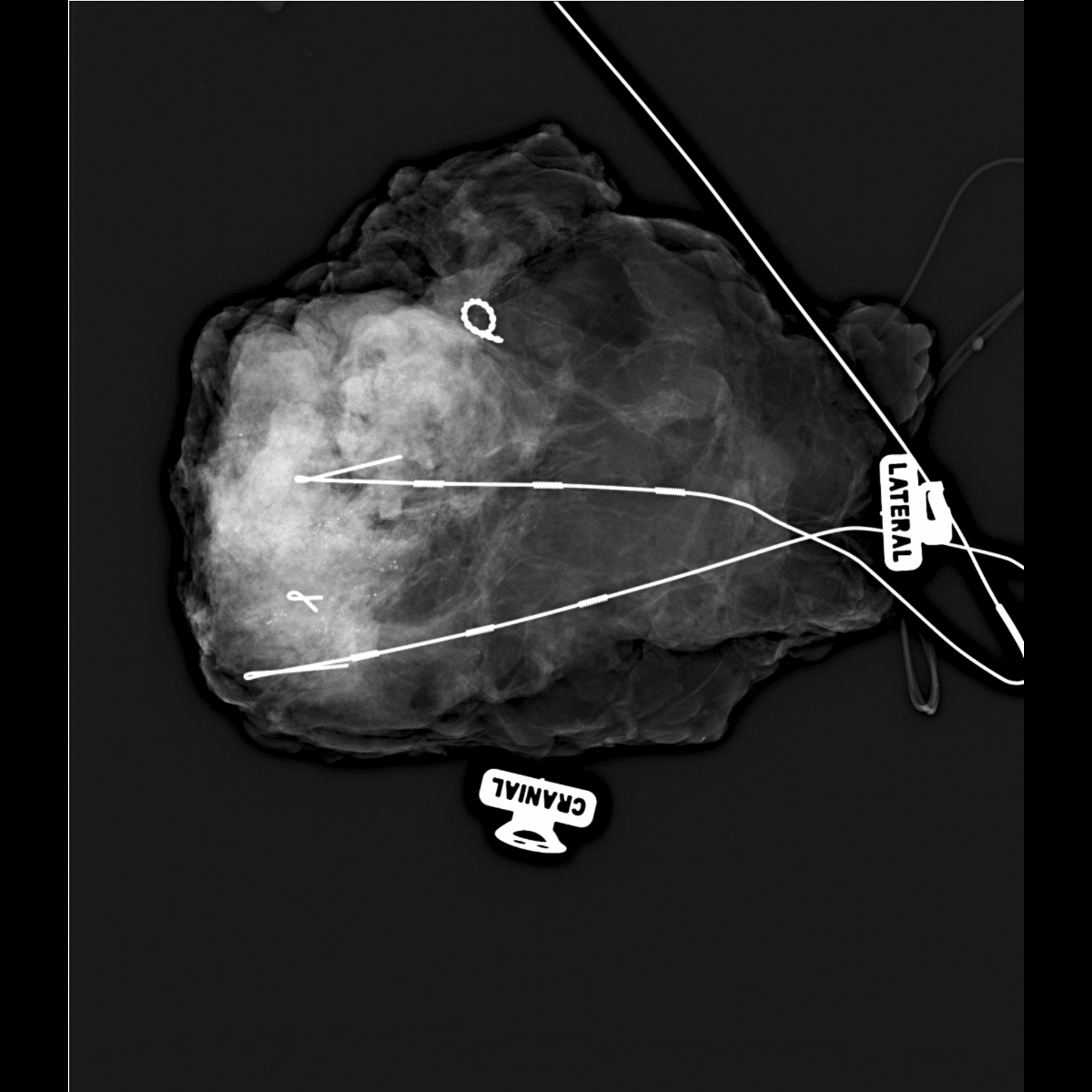} & 
\includegraphics[width=0.24\linewidth]{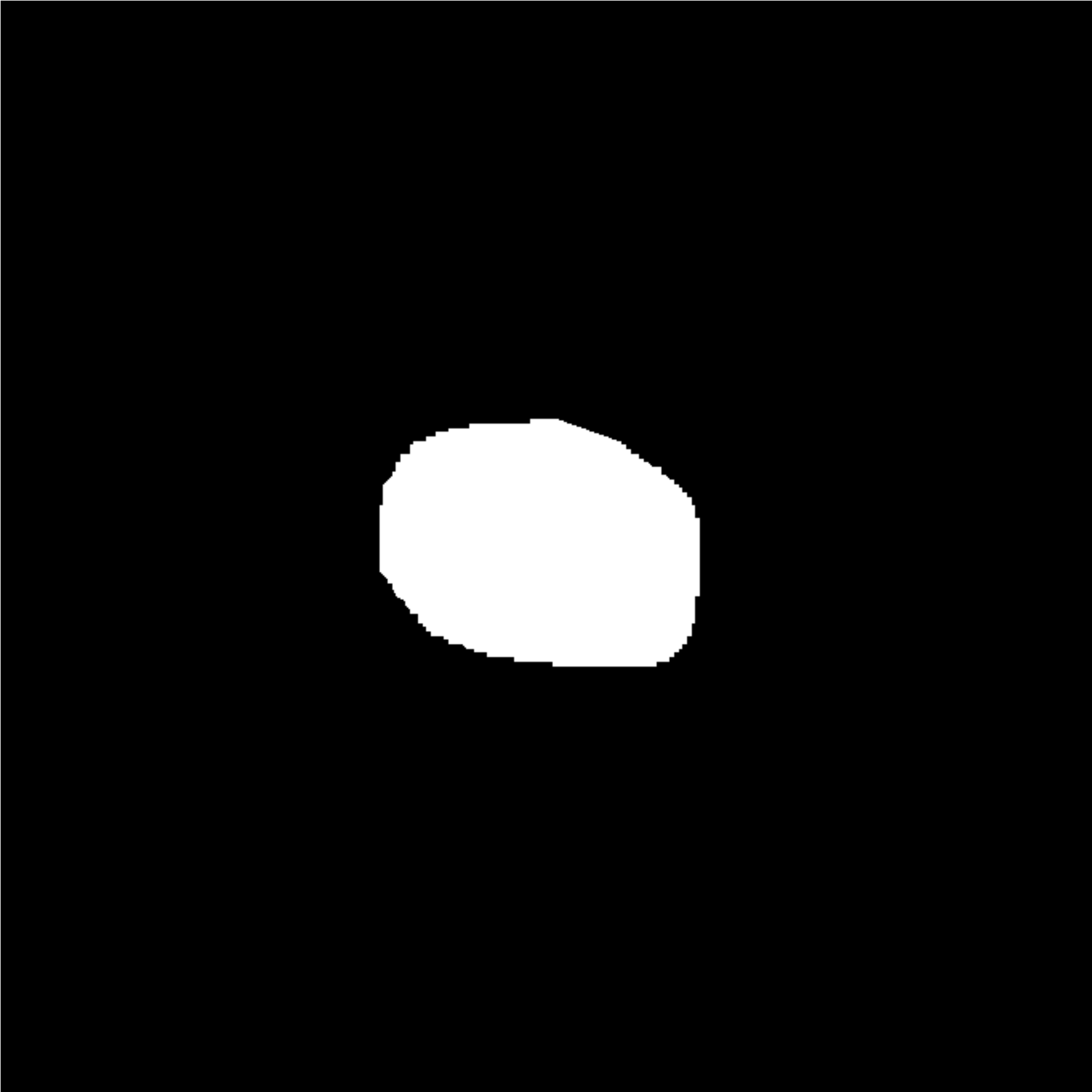} & 
\includegraphics[width=0.24\linewidth]{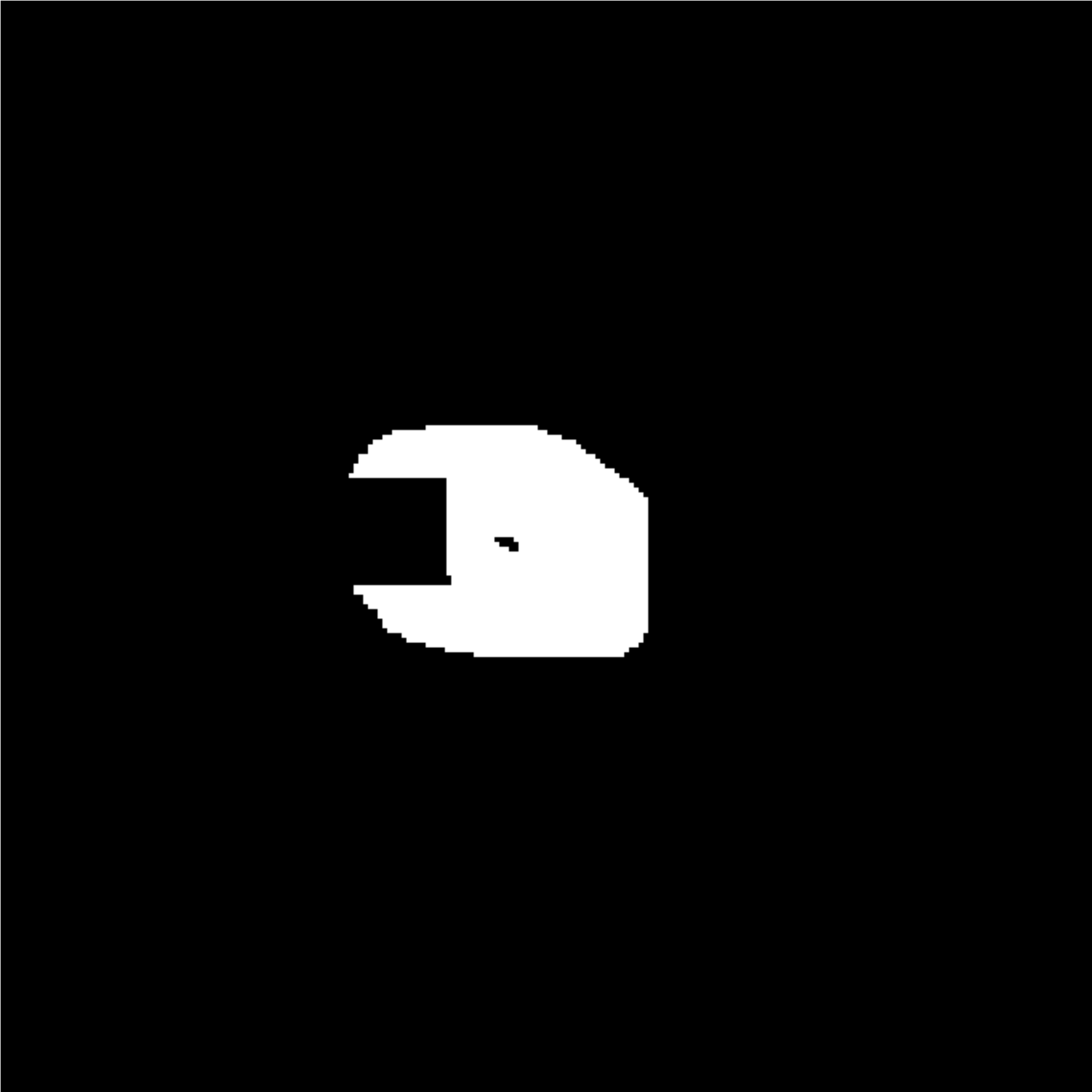} &
\includegraphics[width=0.24\linewidth]{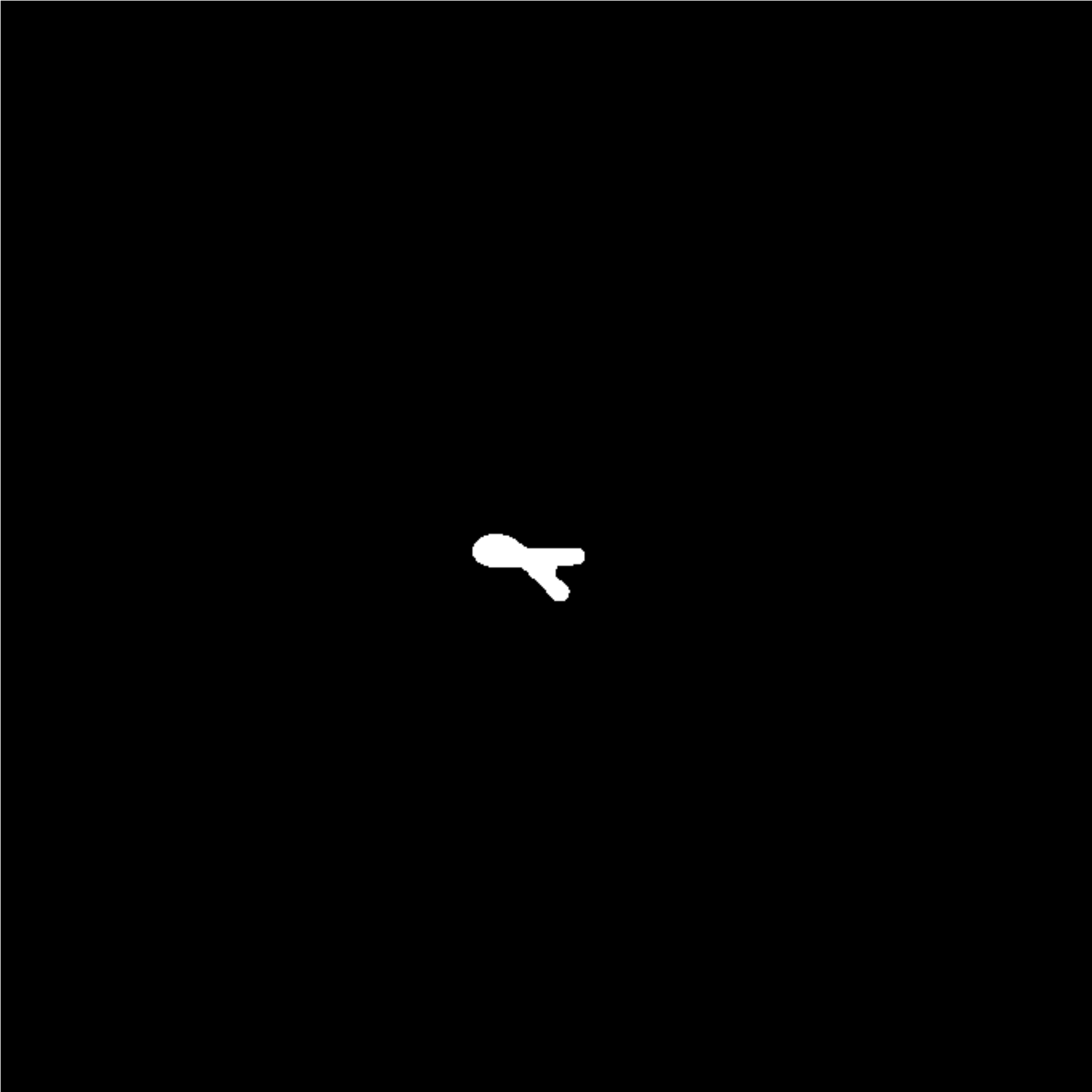} &
\includegraphics[width=0.24\linewidth]{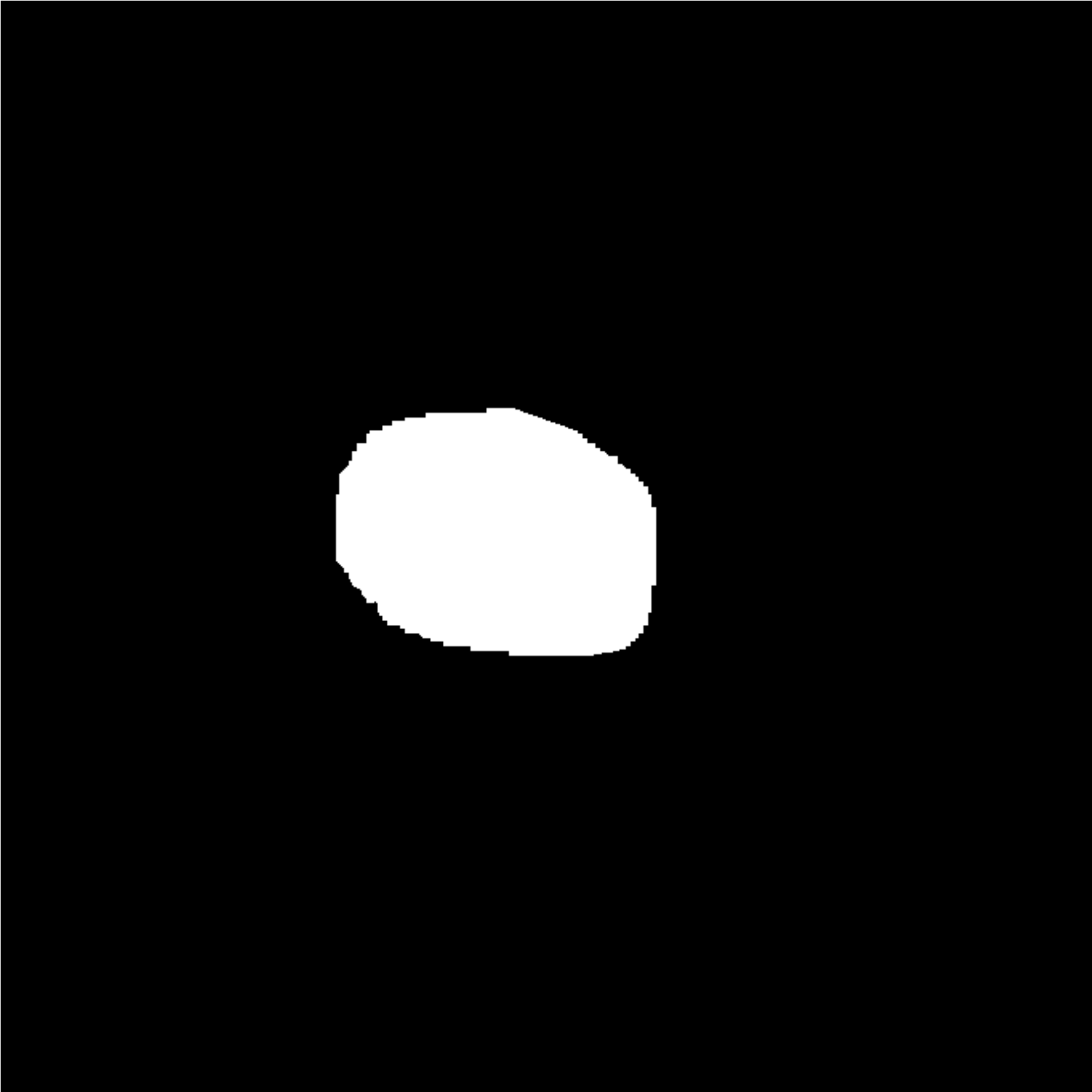} \\
\end{tabular}
}
\caption{Visual comparison of the segmented margins from different SR images. Note that in each case, the SAM 2 refinement process achieves a closer result to the ground truth compared to both the reconstructed and SAM-refined masks.}
\label{fig:segperformance}
\end{figure}

\paragraph{Margin assessment with FFCL-SAM:}

Table~\ref{tab:segresults} reports the segmentation performance for the compared margin mask prediction approaches. The baseline reconstructed masks based on FFCL-based patch-level predictions are relatively poorer, as suggested by all three metrics (DSC, HD, and Accuracy). This is also confirmed from the visualized coarse masks for three different images as shown in Fig.~\ref{fig:segperformance}. The coarse margin masks are passed as prompts and refined via SAM or SAM 2. 

Among zero-shot methods, SAM2 substantially improves over both the baseline and its predecessor, SAM. Specifically, SAM2-zero-shot boosts positive margin segmentation by 25.48\% in DSC, reduces HD by 92.00\%, and improves accuracy by 12.74\%. SAM2 outperforms its predecessor, SAM by 36.55\%, 92.19\%, and 18.28\% in DSC, HD, and accuracy, respectively. The high discrepancy between SAM and SAM 2 results lies in how the prompting is performed. SAM can only be prompted via point or bounding box prompts, which is problematic for segmenting objects with as small and complex a shape as positive margins. 

Furthermore, SAM-based refining techniques are even more effective at segmenting the negative margin (Table~\ref{tab:segresults}). Similar to the positive margin case, SAM 2 outperforms the baseline model by 23.59\%, 94.81\%, and 11.79\%, in terms of DSC, HD, and accuracy, respectively. As expected, SAM 2 outperforms SAM by 15.39\% DSC, 96.97\% HD, and 11.79\% accuracy. Overall, when averaging the results of the positive and negative margin segmentation, SAM 2 outperforms the baseline by 25.54\%, 93.90\%, and 12.27\%, and outperforms SAM by 26.22\%, 95.95\%, and 13.11\%. As seen in Fig.~\ref{fig:segperformance}, the refined margin masks from the SAM 2 model are in better agreement with the ground truth masks. 

However, the highest performance is achieved by the few-shot variant of SAM2, which fine-tunes the model on just five labeled SRs. SAM2-few-shot surpasses all other approaches, achieving 0.9053 DSC, 6.6736 HD, and 95.27\% accuracy for positive margins, with corresponding improvements across negative margin and overall segmentation metrics. Compared to the zero-shot variant, SAM2-few-shot yields a further 2.17\% increase in DSC and 0.96\% gain in accuracy, demonstrating that even limited fine-tuning enhances segmentation quality.

The inference speed of the mask refinement process is 47ms. The improvement in the quality and smoothness of the masks after refinement of the coarse masks justifies the integration of our domain-specific pretraining FFCL and foundational SAM as in FFCL-SAM.

\begin{table}[t]
    \centering
    \caption{Performance comparison of different SAM variants with the baseline reconstructed margin mask for individual positive and negative margins. Segmentation Dice similarity coefficient (DSC), Hausdorff distance (HD), and accuracy scores have been reported by selecting the best FFCL model for initial coarse margin mask generation. Best scores are \textbf{bolded} and Second best are \underline{underlined}.}
    \medskip
    \setlength{\tabcolsep}{4pt}
    \resizebox{\linewidth}{!}{
    \begin{tabular}{lc ccc c ccc c ccc}
    \toprule
     \multirow{2}{*}{Mask} & \phantom{a} & \multicolumn{3}{c}{Positive Margin} & \phantom{a} &  \multicolumn{3}{c}{Negative Margin} \phantom{a} & \multicolumn{3}{c}{Overall}\\
     \cmidrule(lr){3-5} \cmidrule(lr){7-9} \cmidrule(lr){11-13}
     && DSC & HD & Accuracy && DSC & HD & Accuracy && DSC & HD & Accuracy \\
     \midrule
     Reconstructed Mask && 0.6313 & 85.3282 & 0.8157 && 0.6861 & 179.5214 & 0.8431 && 0.6587 & 132.4248 & 0.8294 \\
     SAM-zero shot && 0.5206 & 87.4305 & 0.7603 && 0.7631 & 307.6746 & 0.8816 && 0.6419 & 197.5535 & 0.8210 \\
     SAM2-zero-shot && \underline{0.8861} & \underline{6.8267} & \underline{0.9431} && \underline{0.9220} & \underline{9.3258} & \underline{0.9610} && \underline{0.9041} & \underline{8.0763} & \underline{0.9521} \\
     SAM2-few-shot && \textbf{0.9053} & \textbf{6.6736} & \textbf{0.9527} && \textbf{0.9443} & \textbf{9.0721} & \textbf{0.9722} && \textbf{0.9248} & \textbf{7.8729} & \textbf{0.9625}\\
     \bottomrule
    \end{tabular} }
    \label{tab:segresults}
\end{table}

\section{Conclusions}
\label{sec:conclusions}
We have introduced a novel integration of our innovative self-supervised contrastive pretraining method (FFCL) and a generic foundation model (SAM) for enhanced detection of BCS margins on SR images. The FFCL-based classification model demonstrates the ability to produce accurate patch-level margin labels. The predicted positive and negative patches are filled with all 1s and all 0s, respectively. These initial predictions serve as the foundation for subsequent segmentation tasks, where coarse segmentation masks were constructed from the classified patches and refined using SAM.

FFCL-SAM offers improved classification performance for detecting positive margins compared to baseline models, achieving superior accuracy and AUC values against most of the compared models. Evaluations of the segmentation tasks revealed that SAM 2's zero- and few-shot capabilities showed promise in refining segmentation masks. The findings from this study indicate the potential to reduce the rate of lumpectomy re-excision and improve patient outcomes through the use of AI methods, particularly the integration of domain-specific pre-training with a generic foundation model. 

Despite the promising results of FFCL-SAM, this study has several limitations. \textit{First,} the dataset used in this work is relatively small, which may limit the potential generalizability of the model to a broader range of intraoperative specimens. The dataset imbalance, with fewer positive margin cases, also poses a challenge for training DL models effectively. Additionally, while SAM demonstrates promising performance in refining segmentation masks, its reliance on prompt-based refinement may not always capture subtle variations in margin boundaries, particularly in complex surgical images. Future work will focus on expanding the dataset to improve model robustness, exploring automated prompt generation methods to enhance segmentation accuracy, and validating the approach on diverse patient populations from multiple centers.

\bibliographystyle{ACM-Reference-Format}
\bibliography{sample-base}


\begin{thebibliography}{60}


\ifx \showCODEN    \undefined \def \showCODEN     #1{\unskip}     \fi
\ifx \showISBNx    \undefined \def \showISBNx     #1{\unskip}     \fi
\ifx \showISBNxiii \undefined \def \showISBNxiii  #1{\unskip}     \fi
\ifx \showISSN     \undefined \def \showISSN      #1{\unskip}     \fi
\ifx \showLCCN     \undefined \def \showLCCN      #1{\unskip}     \fi
\ifx \shownote     \undefined \def \shownote      #1{#1}          \fi
\ifx \showarticletitle \undefined \def \showarticletitle #1{#1}   \fi
\ifx \showURL      \undefined \def \showURL       {\relax}        \fi
\providecommand\bibfield[2]{#2}
\providecommand\bibinfo[2]{#2}
\providecommand\natexlab[1]{#1}
\providecommand\showeprint[2][]{arXiv:#2}

\bibitem[Ahamed et~al\mbox{.}(2023)]%
        {ahamed2023ffcl}
\bibfield{author}{\bibinfo{person}{Md~Atik Ahamed}, \bibinfo{person}{Jin Chen}, {and} \bibinfo{person}{Abdullah-Al-Zubaer Imran}.} \bibinfo{year}{2023}\natexlab{}.
\newblock \showarticletitle{{FFCL:} {Forward-Forward} Contrastive Learning for Improved Medical Image Classification}. In \bibinfo{booktitle}{\emph{Medical Imaging with Deep Learning (MIDL), short paper track}}.
\newblock


\bibitem[Ahamed et~al\mbox{.}(2024)]%
        {ahamed2024automatic}
\bibfield{author}{\bibinfo{person}{Md.~Atik Ahamed}, \bibinfo{person}{Braxton McFarland}, \bibinfo{person}{Xiaoqin Wang}, \bibinfo{person}{Jin Chen}, {and} \bibinfo{person}{Abdullah-Al-Zubaer Imran}.} \bibinfo{year}{2024}\natexlab{}.
\newblock \showarticletitle{Automatic detection of breast cancer lumpectomy margin from intraoperative specimen mammography}. In \bibinfo{booktitle}{\emph{17th International Workshop on Breast Imaging (IWBI 2024)}}, Vol.~\bibinfo{volume}{13174}. \bibinfo{publisher}{SPIE}, \bibinfo{pages}{450--454}.
\newblock


\bibitem[Ahuja et~al\mbox{.}(2024)]%
        {ahuja2024comparison}
\bibfield{author}{\bibinfo{person}{Sana Ahuja}, \bibinfo{person}{Priya Yadav}, \bibinfo{person}{Marzieh Fattahi-Darghlou}, {and} \bibinfo{person}{Sufian Zaheer}.} \bibinfo{year}{2024}\natexlab{}.
\newblock \showarticletitle{Comparison of Intraoperative Imprint Cytology versus Frozen Section for Sentinel Lymph Node Evaluation in Breast Cancer. A study along with Systematic Review and Meta-analysis of literature}.
\newblock \bibinfo{journal}{\emph{Asian Pacific Journal of Cancer Prevention: APJCP}} \bibinfo{volume}{25}, \bibinfo{number}{4} (\bibinfo{year}{2024}).
\newblock


\bibitem[Awais et~al\mbox{.}(2025)]%
        {awais2025foundation}
\bibfield{author}{\bibinfo{person}{Muhammad Awais}, \bibinfo{person}{Muzammal Naseer}, \bibinfo{person}{Salman Khan}, \bibinfo{person}{Rao~Muhammad Anwer}, \bibinfo{person}{Hisham Cholakkal}, \bibinfo{person}{Mubarak Shah}, \bibinfo{person}{Ming-Hsuan Yang}, {and} \bibinfo{person}{Fahad~Shahbaz Khan}.} \bibinfo{year}{2025}\natexlab{}.
\newblock \showarticletitle{Foundation Models Defining a New Era in Vision: A Survey and Outlook}.
\newblock \bibinfo{journal}{\emph{IEEE Transactions on Pattern Analysis and Machine Intelligence}} (\bibinfo{year}{2025}).
\newblock


\bibitem[Azad et~al\mbox{.}(2024)]%
        {azad2024medical}
\bibfield{author}{\bibinfo{person}{Reza Azad}, \bibinfo{person}{Ehsan~Khodapanah Aghdam}, \bibinfo{person}{Amelie Rauland}, \bibinfo{person}{Yiwei Jia}, \bibinfo{person}{Atlas~Haddadi Avval}, {and} \bibinfo{person}{Afshin Bozorgpour}.} \bibinfo{year}{2024}\natexlab{}.
\newblock \showarticletitle{Medical Image Segmentation Review: The Success of {U-Net}}.
\newblock \bibinfo{journal}{\emph{IEEE Transactions on Pattern Analysis and Machine Intelligence}} \bibinfo{volume}{46}, \bibinfo{number}{12} (\bibinfo{year}{2024}), \bibinfo{pages}{10076--10095}.
\newblock


\bibitem[Azizi et~al\mbox{.}(2021)]%
        {azizi2021big}
\bibfield{author}{\bibinfo{person}{Shekoofeh Azizi}, \bibinfo{person}{Basil Mustafa}, \bibinfo{person}{Fiona Ryan}, \bibinfo{person}{Zachary Beaver}, \bibinfo{person}{Jan Freyberg}, \bibinfo{person}{Jonathan Deaton}, \bibinfo{person}{Aaron Loh}, \bibinfo{person}{Alan Karthikesalingam}, \bibinfo{person}{Simon Kornblith}, \bibinfo{person}{Ting Chen}, \bibinfo{person}{Vivek Natarajan}, {and} \bibinfo{person}{Mohammad Norouzi}.} \bibinfo{year}{2021}\natexlab{}.
\newblock \showarticletitle{Big self-supervised models advance medical image classification}. In \bibinfo{booktitle}{\emph{Proceedings of the IEEE/CVF International Conference on Computer Vision (ICCV)}}. \bibinfo{pages}{3478--3488}.
\newblock


\bibitem[Cao et~al\mbox{.}(2022)]%
        {cao2022swin}
\bibfield{author}{\bibinfo{person}{Hu Cao}, \bibinfo{person}{Yueyue Wang}, \bibinfo{person}{Joy Chen}, \bibinfo{person}{Dongsheng Jiang}, \bibinfo{person}{Xiaopeng Zhang}, \bibinfo{person}{Qi Tian}, {and} \bibinfo{person}{Manning Wang}.} \bibinfo{year}{2022}\natexlab{}.
\newblock \showarticletitle{{Swin-Unet}: Unet-like pure transformer for medical image segmentation}. In \bibinfo{booktitle}{\emph{European Conference on Computer Vision (ECCV)}}. Springer, \bibinfo{pages}{205--218}.
\newblock


\bibitem[Chakedis et~al\mbox{.}(2022)]%
        {chakedis2022economic}
\bibfield{author}{\bibinfo{person}{Jeffery~M Chakedis}, \bibinfo{person}{Annie Tang}, \bibinfo{person}{Alison Savitz}, \bibinfo{person}{Liisa~L Lyon}, \bibinfo{person}{Patricia~E Palacios}, \bibinfo{person}{Brooke Vuong}, \bibinfo{person}{Maihgan~A Kavanagh}, \bibinfo{person}{Gillian~E Kuehner}, \bibinfo{person}{Sharon~B Chang}, {and} \bibinfo{person}{Permanente Medical Group Breast~Research Collaborative}.} \bibinfo{year}{2022}\natexlab{}.
\newblock \showarticletitle{Economic impact of reducing reexcision rates after breast-conserving surgery in a large, integrated health system}.
\newblock \bibinfo{journal}{\emph{Annals of Surgical Oncology}} \bibinfo{volume}{29}, \bibinfo{number}{10} (\bibinfo{year}{2022}), \bibinfo{pages}{6288--6296}.
\newblock


\bibitem[Chen et~al\mbox{.}(2021)]%
        {chen2021transunet}
\bibfield{author}{\bibinfo{person}{Jieneng Chen}, \bibinfo{person}{Yongyi Lu}, \bibinfo{person}{Qihang Yu}, \bibinfo{person}{Xiangde Luo}, \bibinfo{person}{Ehsan Adeli}, \bibinfo{person}{Yan Wang}, \bibinfo{person}{Le Lu}, \bibinfo{person}{Alan~L Yuille}, {and} \bibinfo{person}{Yuyin Zhou}.} \bibinfo{year}{2021}\natexlab{}.
\newblock \showarticletitle{{TransUNet}: Transformers make strong encoders for medical image segmentation}.
\newblock \bibinfo{journal}{\emph{arXiv preprint arXiv:2102.04306}} (\bibinfo{year}{2021}).
\newblock


\bibitem[Chen et~al\mbox{.}(2024)]%
        {chen2024transunet}
\bibfield{author}{\bibinfo{person}{Jieneng Chen}, \bibinfo{person}{Jieru Mei}, \bibinfo{person}{Xianhang Li}, \bibinfo{person}{Yongyi Lu}, \bibinfo{person}{Qihang Yu}, \bibinfo{person}{Qingyue Wei}, \bibinfo{person}{Xiangde Luo}, \bibinfo{person}{Yutong Xie}, \bibinfo{person}{Ehsan Adeli}, \bibinfo{person}{Yan Wang}, {et~al\mbox{.}}} \bibinfo{year}{2024}\natexlab{}.
\newblock \showarticletitle{{TransUNet}: Rethinking the U-Net architecture design for medical image segmentation through the lens of transformers}.
\newblock \bibinfo{journal}{\emph{Medical Image Analysis}}  \bibinfo{volume}{97} (\bibinfo{year}{2024}).
\newblock


\bibitem[Chen et~al\mbox{.}(2023)]%
        {chen2023analysis}
\bibfield{author}{\bibinfo{person}{Kevin~A Chen}, \bibinfo{person}{Kathryn~E Kirchoff}, \bibinfo{person}{Logan~R Butler}, \bibinfo{person}{Alexa~D Holloway}, \bibinfo{person}{Muneera~R Kapadia}, \bibinfo{person}{Cherie~M Kuzmiak}, \bibinfo{person}{Stephanie~M Downs-Canner}, \bibinfo{person}{Phillip~M Spanheimer}, \bibinfo{person}{Kristalyn~K Gallagher}, {and} \bibinfo{person}{Shawn~M Gomez}.} \bibinfo{year}{2023}\natexlab{}.
\newblock \showarticletitle{Analysis of Specimen Mammography with Artificial Intelligence to Predict Margin Status}.
\newblock \bibinfo{journal}{\emph{Annals of Surgical Oncology}} \bibinfo{volume}{30}, \bibinfo{number}{12} (\bibinfo{year}{2023}), \bibinfo{pages}{7107--7115}.
\newblock


\bibitem[Chen et~al\mbox{.}(2020)]%
        {chen2020a}
\bibfield{author}{\bibinfo{person}{Ting Chen}, \bibinfo{person}{Simon Kornblith}, \bibinfo{person}{Mohammad Norouzi}, {and} \bibinfo{person}{Geoffrey Hinton}.} \bibinfo{year}{2020}\natexlab{}.
\newblock \showarticletitle{A simple framework for contrastive learning of visual representations}. In \bibinfo{booktitle}{\emph{Proceedings of the 37th International Conference on Machine Learning}}, Vol.~\bibinfo{volume}{119}. \bibinfo{publisher}{PMLR}, \bibinfo{pages}{1597--1607}.
\newblock


\bibitem[{\c{C}}orbac{\i}o{\u{g}}lu and Aksel(2023)]%
        {ccorbaciouglu2023receiver}
\bibfield{author}{\bibinfo{person}{{\c{S}}eref~Kerem {\c{C}}orbac{\i}o{\u{g}}lu} {and} \bibinfo{person}{G{\"o}khan Aksel}.} \bibinfo{year}{2023}\natexlab{}.
\newblock \showarticletitle{Receiver operating characteristic curve analysis in diagnostic accuracy studies: A guide to interpreting the area under the curve value}.
\newblock \bibinfo{journal}{\emph{Turkish journal of emergency medicine}} \bibinfo{volume}{23}, \bibinfo{number}{4} (\bibinfo{year}{2023}), \bibinfo{pages}{195--198}.
\newblock


\bibitem[David et~al\mbox{.}(2023)]%
        {david2023situ}
\bibfield{author}{\bibinfo{person}{Sandryne David}, \bibinfo{person}{Trang Tran}, \bibinfo{person}{Fr{\'e}d{\'e}rick Dallaire}, \bibinfo{person}{Guillaume Sheehy}, \bibinfo{person}{Feryel Azzi}, \bibinfo{person}{Dominique Trudel}, \bibinfo{person}{Francine Tremblay}, \bibinfo{person}{Atilla Omeroglu}, \bibinfo{person}{Fr{\'e}d{\'e}ric Leblond}, {and} \bibinfo{person}{Sarkis Meterissian}.} \bibinfo{year}{2023}\natexlab{}.
\newblock \showarticletitle{In situ {Raman} spectroscopy and machine learning unveil biomolecular alterations in invasive breast cancer}.
\newblock \bibinfo{journal}{\emph{Journal of Biomedical Optics}} \bibinfo{volume}{28}, \bibinfo{number}{3} (\bibinfo{year}{2023}), \bibinfo{pages}{036009--036009}.
\newblock


\bibitem[Dosovitskiy(2020)]%
        {dosovitskiy2020image}
\bibfield{author}{\bibinfo{person}{Alexey Dosovitskiy}.} \bibinfo{year}{2020}\natexlab{}.
\newblock \showarticletitle{An image is worth 16x16 words: Transformers for image recognition at scale}.
\newblock \bibinfo{journal}{\emph{arXiv preprint arXiv:2010.11929}} (\bibinfo{year}{2020}).
\newblock


\bibitem[Dowling et~al\mbox{.}(2024)]%
        {dowling2024diagnostic}
\bibfield{author}{\bibinfo{person}{Gavin~P Dowling}, \bibinfo{person}{Cian~M Hehir}, \bibinfo{person}{Gordon~R Daly}, \bibinfo{person}{Sandra Hembrecht}, \bibinfo{person}{Stephen Keelan}, \bibinfo{person}{Katie Giblin}, \bibinfo{person}{Maen~M Alrawashdeh}, \bibinfo{person}{Fiona Boland}, {and} \bibinfo{person}{Arnold~DK Hill}.} \bibinfo{year}{2024}\natexlab{}.
\newblock \showarticletitle{Diagnostic accuracy of intraoperative methods for margin assessment in breast cancer surgery: A systematic review \& meta-analysis}.
\newblock \bibinfo{journal}{\emph{The Breast}} (\bibinfo{year}{2024}).
\newblock


\bibitem[D’Alfonso et~al\mbox{.}(2021)]%
        {d2021multi}
\bibfield{author}{\bibinfo{person}{Timothy~M D’Alfonso}, \bibinfo{person}{David~Joon Ho}, \bibinfo{person}{Matthew~G Hanna}, \bibinfo{person}{Anne Grabenstetter}, \bibinfo{person}{Dig Vijay~Kumar Yarlagadda}, \bibinfo{person}{Luke Geneslaw}, \bibinfo{person}{Peter Ntiamoah}, \bibinfo{person}{Thomas~J Fuchs}, {and} \bibinfo{person}{Lee~K Tan}.} \bibinfo{year}{2021}\natexlab{}.
\newblock \showarticletitle{Multi-magnification-based machine learning as an ancillary tool for the pathologic assessment of shaved margins for breast carcinoma lumpectomy specimens}.
\newblock \bibinfo{journal}{\emph{Modern Pathology}} \bibinfo{volume}{34}, \bibinfo{number}{8} (\bibinfo{year}{2021}), \bibinfo{pages}{1487--1494}.
\newblock


\bibitem[Ericsson et~al\mbox{.}(2022)]%
        {ericsson2022self}
\bibfield{author}{\bibinfo{person}{Linus Ericsson}, \bibinfo{person}{Henry Gouk}, \bibinfo{person}{Chen~Change Loy}, {and} \bibinfo{person}{Timothy~M. Hospedales}.} \bibinfo{year}{2022}\natexlab{}.
\newblock \showarticletitle{Self-supervised representation learning: Introduction, advances, and challenges}.
\newblock \bibinfo{journal}{\emph{IEEE Signal Processing Magazine}} \bibinfo{volume}{39}, \bibinfo{number}{3} (\bibinfo{year}{2022}), \bibinfo{pages}{42--62}.
\newblock


\bibitem[Esbona et~al\mbox{.}(2012)]%
        {esbona2012intraoperative}
\bibfield{author}{\bibinfo{person}{Karla Esbona}, \bibinfo{person}{Zhanhai Li}, {and} \bibinfo{person}{Lee~G Wilke}.} \bibinfo{year}{2012}\natexlab{}.
\newblock \showarticletitle{Intraoperative imprint cytology and frozen section pathology for margin assessment in breast conservation surgery: a systematic review}.
\newblock \bibinfo{journal}{\emph{Annals of Surgical Oncology}}  \bibinfo{volume}{19} (\bibinfo{year}{2012}), \bibinfo{pages}{3236--3245}.
\newblock


\bibitem[Garza et~al\mbox{.}(2024)]%
        {garza2024intraoperative}
\bibfield{author}{\bibinfo{person}{Kyana~Y Garza}, \bibinfo{person}{Mary~E King}, \bibinfo{person}{Chandandeep Nagi}, \bibinfo{person}{Rachel~J DeHoog}, \bibinfo{person}{Jialing Zhang}, \bibinfo{person}{Marta Sans}, \bibinfo{person}{Anna Krieger}, \bibinfo{person}{Clara~L Feider}, \bibinfo{person}{Alena~V Bensussan}, \bibinfo{person}{Michael~F Keating}, {et~al\mbox{.}}} \bibinfo{year}{2024}\natexlab{}.
\newblock \showarticletitle{Intraoperative Evaluation of Breast Tissues During Breast Cancer Operations Using the MasSpec Pen}.
\newblock \bibinfo{journal}{\emph{JAMA Network Open}} \bibinfo{volume}{7}, \bibinfo{number}{3} (\bibinfo{year}{2024}), \bibinfo{pages}{e242684--e242684}.
\newblock


\bibitem[Grill et~al\mbox{.}(2020)]%
        {grill2020bootstrap}
\bibfield{author}{\bibinfo{person}{Jean-Bastien Grill}, \bibinfo{person}{Florian Strub}, \bibinfo{person}{Florent Altch\'{e}}, \bibinfo{person}{Corentin Tallec}, \bibinfo{person}{Pierre Richemond}, \bibinfo{person}{Elena Buchatskaya}, \bibinfo{person}{Carl Doersch}, \bibinfo{person}{Bernardo Avila~Pires}, \bibinfo{person}{Zhaohan Guo}, \bibinfo{person}{Mohammad Gheshlaghi~Azar}, \bibinfo{person}{Bilal Piot}, \bibinfo{person}{koray kavukcuoglu}, \bibinfo{person}{Remi Munos}, {and} \bibinfo{person}{Michal Valko}.} \bibinfo{year}{2020}\natexlab{}.
\newblock \showarticletitle{{Bootstrap Your Own Latent} - A new approach to self-supervised learning}. In \bibinfo{booktitle}{\emph{Advances in Neural Information Processing Systems}}, Vol.~\bibinfo{volume}{33}. \bibinfo{publisher}{Curran Associates, Inc.}, \bibinfo{pages}{21271--21284}.
\newblock


\bibitem[He et~al\mbox{.}(2020)]%
        {he2020momentum}
\bibfield{author}{\bibinfo{person}{Kaiming He}, \bibinfo{person}{Haoqi Fan}, \bibinfo{person}{Yuxin Wu}, \bibinfo{person}{Saining Xie}, {and} \bibinfo{person}{Ross Girshick}.} \bibinfo{year}{2020}\natexlab{}.
\newblock \showarticletitle{{Momentum Contrast} for unsupervised visual representation learning}. In \bibinfo{booktitle}{\emph{Proceedings of the IEEE/CVF Conference on Computer Vision and Pattern Recognition (CVPR)}}. \bibinfo{pages}{9729--9738}.
\newblock


\bibitem[He et~al\mbox{.}(2016)]%
        {he2016deep}
\bibfield{author}{\bibinfo{person}{Kaiming He}, \bibinfo{person}{Xiangyu Zhang}, \bibinfo{person}{Shaoqing Ren}, {and} \bibinfo{person}{Jian Sun}.} \bibinfo{year}{2016}\natexlab{}.
\newblock \showarticletitle{Deep Residual Learning for Image Recognition}. In \bibinfo{booktitle}{\emph{Proceedings of the IEEE/CVF Conference on Computer Vision and Pattern Recognition (CVPR)}}. \bibinfo{pages}{770--778}.
\newblock


\bibitem[Hinton(2022)]%
        {hinton2022forward}
\bibfield{author}{\bibinfo{person}{Geoffrey Hinton}.} \bibinfo{year}{2022}\natexlab{}.
\newblock \showarticletitle{The {Forward-Forward Algorithm}: Some preliminary investigations}.
\newblock \bibinfo{journal}{\emph{arXiv preprint arXiv:2212.13345}} (\bibinfo{year}{2022}).
\newblock


\bibitem[Ho et~al\mbox{.}(2020)]%
        {ho2020deep}
\bibfield{author}{\bibinfo{person}{David~Joon Ho}, \bibinfo{person}{Narasimhan~P Agaram}, \bibinfo{person}{Peter~J Sch{\"u}ffler}, \bibinfo{person}{Chad~M Vanderbilt}, \bibinfo{person}{Marc-Henri Jean}, \bibinfo{person}{Meera~R Hameed}, {and} \bibinfo{person}{Thomas~J Fuchs}.} \bibinfo{year}{2020}\natexlab{}.
\newblock \showarticletitle{Deep interactive learning: An efficient labeling approach for deep learning-based osteosarcoma treatment response assessment}. In \bibinfo{booktitle}{\emph{Medical Image Computing and Computer Assisted Intervention--MICCAI 2020: 23rd International Conference, Lima, Peru, October 4--8, 2020, Proceedings, Part V 23}}. Springer, \bibinfo{pages}{540--549}.
\newblock


\bibitem[Hu et~al\mbox{.}(2024)]%
        {hu2024a}
\bibfield{author}{\bibinfo{person}{Hiagen Hu}, \bibinfo{person}{Xiaoyuan Wang}, \bibinfo{person}{Yan Zhang}, \bibinfo{person}{Qi Chen}, {and} \bibinfo{person}{Qiu Guan}.} \bibinfo{year}{2024}\natexlab{}.
\newblock \showarticletitle{A comprehensive survey on contrastive learning}.
\newblock \bibinfo{journal}{\emph{Neurocomputing}}  \bibinfo{volume}{610} (\bibinfo{year}{2024}).
\newblock


\bibitem[Huang et~al\mbox{.}(2024)]%
        {huang2024enhancing}
\bibfield{author}{\bibinfo{person}{Weijian Huang}, \bibinfo{person}{Cheng Li}, \bibinfo{person}{Hong-Yu Zhou}, \bibinfo{person}{Hao Yang}, \bibinfo{person}{Jiarun Liu}, \bibinfo{person}{Yong Liang}, \bibinfo{person}{Hairong Zheng}, \bibinfo{person}{Shaoting Zhang}, {and} \bibinfo{person}{Shanshan Wang}.} \bibinfo{year}{2024}\natexlab{}.
\newblock \showarticletitle{Enhancing representation in radiography-reports foundation model: a granular alignment algorithm using masked contrastive learning}.
\newblock \bibinfo{journal}{\emph{Nature Communications}}  \bibinfo{volume}{15} (\bibinfo{year}{2024}).
\newblock


\bibitem[Katipamula et~al\mbox{.}(2009)]%
        {katipamula2009trends}
\bibfield{author}{\bibinfo{person}{Rajini Katipamula}, \bibinfo{person}{Amy~C Degnim}, \bibinfo{person}{Tanya Hoskin}, \bibinfo{person}{Judy~C Boughey}, \bibinfo{person}{Charles Loprinzi}, \bibinfo{person}{Clive~S Grant}, \bibinfo{person}{Kathleen~R Brandt}, \bibinfo{person}{Sandhya Pruthi}, \bibinfo{person}{Christopher~G Chute}, \bibinfo{person}{Janet~E Olson}, {et~al\mbox{.}}} \bibinfo{year}{2009}\natexlab{}.
\newblock \showarticletitle{Trends in mastectomy rates at the {Mayo Clinic Rochester}: Effect of surgical year and preoperative magnetic resonance imaging}.
\newblock \bibinfo{journal}{\emph{Journal of Clinical Oncology}} \bibinfo{volume}{27}, \bibinfo{number}{25} (\bibinfo{year}{2009}), \bibinfo{pages}{4082}.
\newblock


\bibitem[Kirillov et~al\mbox{.}(2023)]%
        {kirillov2023segment}
\bibfield{author}{\bibinfo{person}{Alexander Kirillov}, \bibinfo{person}{Eric Mintun}, \bibinfo{person}{Nikhila Ravi}, \bibinfo{person}{Hanzi Mao}, \bibinfo{person}{Chloe Rolland}, \bibinfo{person}{Laura Gustafson}, \bibinfo{person}{Tete Xiao}, \bibinfo{person}{Spencer Whitehead}, \bibinfo{person}{Alexander~C Berg}, \bibinfo{person}{Wan-Yen Lo}, {et~al\mbox{.}}} \bibinfo{year}{2023}\natexlab{}.
\newblock \showarticletitle{Segment anything}. In \bibinfo{booktitle}{\emph{Proceedings of the IEEE/CVF International Conference on Computer Vision (ICCV)}}. \bibinfo{pages}{4015--4026}.
\newblock


\bibitem[Lautner et~al\mbox{.}(2015)]%
        {lautner2015disparities}
\bibfield{author}{\bibinfo{person}{Meeghan Lautner}, \bibinfo{person}{Heather Lin}, \bibinfo{person}{Yu Shen}, \bibinfo{person}{Catherine Parker}, \bibinfo{person}{Henry Kuerer}, \bibinfo{person}{Simona Shaitelman}, \bibinfo{person}{Gildy Babiera}, {and} \bibinfo{person}{Isabelle Bedrosian}.} \bibinfo{year}{2015}\natexlab{}.
\newblock \showarticletitle{Disparities in the use of breast-conserving therapy among patients with early-stage breast cancer}.
\newblock \bibinfo{journal}{\emph{JAMA Surgery}} \bibinfo{volume}{150}, \bibinfo{number}{8} (\bibinfo{year}{2015}), \bibinfo{pages}{778--786}.
\newblock


\bibitem[Le-Khac et~al\mbox{.}(2020)]%
        {lekhac2020contrastive}
\bibfield{author}{\bibinfo{person}{Phuc~H. Le-Khac}, \bibinfo{person}{Graham Healy}, {and} \bibinfo{person}{Alan~F. Smeaton}.} \bibinfo{year}{2020}\natexlab{}.
\newblock \showarticletitle{Contrastive representation learning: A framework and review}.
\newblock \bibinfo{journal}{\emph{IEEE Access}}  \bibinfo{volume}{8} (\bibinfo{year}{2020}), \bibinfo{pages}{193907--193934}.
\newblock


\bibitem[Li et~al\mbox{.}(2024)]%
        {li2024mresunet}
\bibfield{author}{\bibinfo{person}{Pengcheng Li}, \bibinfo{person}{Li Zhihao}, \bibinfo{person}{Zijian Wang}, \bibinfo{person}{Li Chaoxiang}, {and} \bibinfo{person}{Monan Wang}.} \bibinfo{year}{2024}\natexlab{}.
\newblock \showarticletitle{{mResU-Net}: multi-scale residual {U-Net}-based brain tumor segmentation from multimodal {MRI}}.
\newblock \bibinfo{journal}{\emph{Medical \& Biological Engineering \& Computing}} \bibinfo{volume}{62}, \bibinfo{number}{3} (\bibinfo{year}{2024}), \bibinfo{pages}{641--651}.
\newblock


\bibitem[Li et~al\mbox{.}(2021)]%
        {li2021domain}
\bibfield{author}{\bibinfo{person}{Zheren Li}, \bibinfo{person}{Zhiming Cui}, \bibinfo{person}{Sheng Wang}, \bibinfo{person}{Yuji Qi}, \bibinfo{person}{Xi Ouyang}, \bibinfo{person}{Qitian Chen}, \bibinfo{person}{Yuezhi Yang}, \bibinfo{person}{Zhong Xue}, \bibinfo{person}{Dinggang Shen}, {and} \bibinfo{person}{Jie-Zhi Cheng}.} \bibinfo{year}{2021}\natexlab{}.
\newblock \showarticletitle{Domain generalization for mammography detection via multi-style and multi-view contrastive learning}. In \bibinfo{booktitle}{\emph{Medical Image Computing and Computer Assisted Intervention -- MICCAI 2021}}. \bibinfo{publisher}{Springer International Publishing}, \bibinfo{pages}{98--108}.
\newblock


\bibitem[Lin et~al\mbox{.}(2017)]%
        {lin2017focal}
\bibfield{author}{\bibinfo{person}{Tsung-Yi Lin}, \bibinfo{person}{Priya Goyal}, \bibinfo{person}{Ross Girshick}, \bibinfo{person}{Kaiming He}, {and} \bibinfo{person}{Piotr Doll{\'a}r}.} \bibinfo{year}{2017}\natexlab{}.
\newblock \showarticletitle{Focal loss for dense object detection}. In \bibinfo{booktitle}{\emph{Proceedings of the IEEE international conference on computer vision}}. \bibinfo{pages}{2980--2988}.
\newblock


\bibitem[Liu et~al\mbox{.}(2021)]%
        {liu2021swin}
\bibfield{author}{\bibinfo{person}{Ze Liu}, \bibinfo{person}{Yutong Lin}, \bibinfo{person}{Yue Cao}, \bibinfo{person}{Han Hu}, \bibinfo{person}{Yixuan Wei}, \bibinfo{person}{Zheng Zhang}, \bibinfo{person}{Stephen Lin}, {and} \bibinfo{person}{Baining Guo}.} \bibinfo{year}{2021}\natexlab{}.
\newblock \showarticletitle{Swin transformer: Hierarchical vision transformer using shifted windows}. In \bibinfo{booktitle}{\emph{Proceedings of the IEEE/CVF International Conference on Computer Vision (ICCV)}}. \bibinfo{pages}{10012--10022}.
\newblock


\bibitem[Liu et~al\mbox{.}(2022)]%
        {liu2022convnet}
\bibfield{author}{\bibinfo{person}{Zhuang Liu}, \bibinfo{person}{Hanzi Mao}, \bibinfo{person}{Chao-Yuan Wu}, \bibinfo{person}{Christoph Feichtenhofer}, \bibinfo{person}{Trevor Darrell}, {and} \bibinfo{person}{Saining Xie}.} \bibinfo{year}{2022}\natexlab{}.
\newblock \showarticletitle{A {ConvNet} for the 2020s}. In \bibinfo{booktitle}{\emph{Proceedings of the IEEE/CVF Conference on Computer Vision and Pattern Recognition (CVPR)}}. \bibinfo{pages}{11976--11986}.
\newblock


\bibitem[Lu et~al\mbox{.}(2022)]%
        {lu2022automated}
\bibfield{author}{\bibinfo{person}{Tongtong Lu}, \bibinfo{person}{Julie~M Jorns}, \bibinfo{person}{Dong~Hye Ye}, \bibinfo{person}{Mollie Patton}, \bibinfo{person}{Renee Fisher}, \bibinfo{person}{Amanda Emmrich}, \bibinfo{person}{Taly~Gilat Schmidt}, \bibinfo{person}{Tina Yen}, {and} \bibinfo{person}{Bing Yu}.} \bibinfo{year}{2022}\natexlab{}.
\newblock \showarticletitle{Automated assessment of breast margins in deep ultraviolet fluorescence images using texture analysis}.
\newblock \bibinfo{journal}{\emph{Biomedical Optics Express}} \bibinfo{volume}{13}, \bibinfo{number}{9} (\bibinfo{year}{2022}), \bibinfo{pages}{5015--5034}.
\newblock


\bibitem[Ma and Fei(2021)]%
        {ma2021comprehensive}
\bibfield{author}{\bibinfo{person}{Ling Ma} {and} \bibinfo{person}{Baowei Fei}.} \bibinfo{year}{2021}\natexlab{}.
\newblock \showarticletitle{Comprehensive review of surgical microscopes: Technology development and medical applications}.
\newblock \bibinfo{journal}{\emph{Journal of Biomedical Optics}} \bibinfo{volume}{26}, \bibinfo{number}{1} (\bibinfo{year}{2021}), \bibinfo{pages}{010901--010901}.
\newblock


\bibitem[Moran et~al\mbox{.}(2014)]%
        {moran2014society}
\bibfield{author}{\bibinfo{person}{Meena~S Moran}, \bibinfo{person}{Stuart~J Schnitt}, \bibinfo{person}{Armando~E Giuliano}, \bibinfo{person}{Jay~R Harris}, \bibinfo{person}{Seema~A Khan}, \bibinfo{person}{Janet Horton}, \bibinfo{person}{Suzanne Klimberg}, \bibinfo{person}{Mariana Chavez-MacGregor}, \bibinfo{person}{Gary Freedman}, \bibinfo{person}{Nehmat Houssami}, {et~al\mbox{.}}} \bibinfo{year}{2014}\natexlab{}.
\newblock \showarticletitle{Society of Surgical Oncology--American Society for Radiation Oncology consensus guideline on margins for breast-conserving surgery with whole-breast irradiation in stages I and II invasive breast cancer}.
\newblock \bibinfo{journal}{\emph{International Journal of Radiation Oncology* Biology* Physics}} \bibinfo{volume}{88}, \bibinfo{number}{3} (\bibinfo{year}{2014}), \bibinfo{pages}{553--564}.
\newblock


\bibitem[Müller and Kainz(2024)]%
        {muller2024resource}
\bibfield{author}{\bibinfo{person}{Johanna~P. Müller} {and} \bibinfo{person}{Bernhard Kainz}.} \bibinfo{year}{2024}\natexlab{}.
\newblock \showarticletitle{Resource-efficient medical image analysis with self-adapting {Forward-Forward} networks}. In \bibinfo{booktitle}{\emph{Machine Learning in Medical Imaging}}. \bibinfo{publisher}{Springer Nature Switzerland}, \bibinfo{pages}{180--190}.
\newblock


\bibitem[Pradipta et~al\mbox{.}(2020)]%
        {pradipta2020emerging}
\bibfield{author}{\bibinfo{person}{Ambara~R Pradipta}, \bibinfo{person}{Tomonori Tanei}, \bibinfo{person}{Koji Morimoto}, \bibinfo{person}{Kenzo Shimazu}, \bibinfo{person}{Shinzaburo Noguchi}, {and} \bibinfo{person}{Katsunori Tanaka}.} \bibinfo{year}{2020}\natexlab{}.
\newblock \showarticletitle{Emerging technologies for real-time intraoperative margin assessment in future breast-conserving surgery}.
\newblock \bibinfo{journal}{\emph{Advanced Science}} \bibinfo{volume}{7}, \bibinfo{number}{9} (\bibinfo{year}{2020}), \bibinfo{pages}{1901519}.
\newblock


\bibitem[Pramanik et~al\mbox{.}(2024)]%
        {pramanik2024daunet}
\bibfield{author}{\bibinfo{person}{Payel Pramanik} {et~al\mbox{.}}} \bibinfo{year}{2024}\natexlab{}.
\newblock \showarticletitle{{DAU-Net}: Dual attention-aided {U-Net} for segmenting tumor in breast ultrasound images}.
\newblock \bibinfo{journal}{\emph{PLoS ONE}} \bibinfo{volume}{19}, \bibinfo{number}{5} (\bibinfo{year}{2024}).
\newblock


\bibitem[Pribadi et~al\mbox{.}(2024)]%
        {pribadi2024optimization}
\bibfield{author}{\bibinfo{person}{Salma~Nurhaliza Pribadi}, \bibinfo{person}{Igi Ardiyanto}, {and} \bibinfo{person}{Nahar Taufiq}.} \bibinfo{year}{2024}\natexlab{}.
\newblock \showarticletitle{Optimization of segmentation for true lumen, false lumen, and false lumen thrombus in type B aortic dissection using the {2D} fine-tuning {U-Net} method}. In \bibinfo{booktitle}{\emph{Sixth International Conference on Image, Video Processing, and Artificial Intelligence (IVPAI 2024)}}, Vol.~\bibinfo{volume}{13225}. \bibinfo{publisher}{SPIE}, \bibinfo{pages}{210--218}.
\newblock


\bibitem[Rahman et~al\mbox{.}(2024)]%
        {rahman2024pp}
\bibfield{author}{\bibinfo{person}{Md~Mostafijur Rahman}, \bibinfo{person}{Mustafa Munir}, \bibinfo{person}{Debesh Jha}, \bibinfo{person}{Ulas Bagci}, {and} \bibinfo{person}{Radu Marculescu}.} \bibinfo{year}{2024}\natexlab{}.
\newblock \showarticletitle{{PP-SAM}: Perturbed Prompts for Robust Adaption of Segment Anything Model for Polyp Segmentation}. In \bibinfo{booktitle}{\emph{Proceedings of the IEEE/CVF Conference on Computer Vision and Pattern Recognition (CVPR)}}. \bibinfo{pages}{4989--4995}.
\newblock


\bibitem[Rakha et~al\mbox{.}(2024)]%
        {rakha2024revisiting}
\bibfield{author}{\bibinfo{person}{Emad~A Rakha}, \bibinfo{person}{Cecily Quinn}, \bibinfo{person}{Yazan~A Masannat}, \bibinfo{person}{Andrew~HS Lee}, \bibinfo{person}{Puay~Hoon Tan}, \bibinfo{person}{Andreas Karakatsanis}, \bibinfo{person}{Zoltan~Tamas Matrai}, \bibinfo{person}{Salman Husain~M Al~Shaibani}, \bibinfo{person}{Salahddin~A Gehani}, \bibinfo{person}{Abeer Shaaban}, {et~al\mbox{.}}} \bibinfo{year}{2024}\natexlab{}.
\newblock \showarticletitle{Revisiting surgical margins for invasive breast cancer patients treated with breast conservation therapy-Evidence for adopting a 1 mm negative width}.
\newblock \bibinfo{journal}{\emph{European Journal of Surgical Oncology}} \bibinfo{volume}{50}, \bibinfo{number}{10} (\bibinfo{year}{2024}), \bibinfo{pages}{108573}.
\newblock


\bibitem[Ravi et~al\mbox{.}(2024)]%
        {ravi2024sam}
\bibfield{author}{\bibinfo{person}{Nikhila Ravi}, \bibinfo{person}{Valentin Gabeur}, \bibinfo{person}{Yuan-Ting Hu}, \bibinfo{person}{Ronghang Hu}, \bibinfo{person}{Chaitanya Ryali}, \bibinfo{person}{Tengyu Ma}, \bibinfo{person}{Haitham Khedr}, \bibinfo{person}{Roman R{\"a}dle}, \bibinfo{person}{Chloe Rolland}, \bibinfo{person}{Laura Gustafson}, {et~al\mbox{.}}} \bibinfo{year}{2024}\natexlab{}.
\newblock \showarticletitle{{SAM 2}: Segment anything in images and videos}.
\newblock \bibinfo{journal}{\emph{arXiv preprint arXiv:2408.00714}} (\bibinfo{year}{2024}).
\newblock


\bibitem[Reyes-Angulo and Paheding(2023)]%
        {angulo2023the}
\bibfield{author}{\bibinfo{person}{Abel Reyes-Angulo} {and} \bibinfo{person}{Sidike Paheding}.} \bibinfo{year}{2023}\natexlab{}.
\newblock \showarticletitle{{The Forward-Forward Algorithm} as a feature extractor for skin lesion classification: A preliminary study}.
\newblock \bibinfo{journal}{\emph{arXiv preprint arXiv:2307.00617}} (\bibinfo{year}{2023}).
\newblock


\bibitem[Ronneberger et~al\mbox{.}(2015)]%
        {ronneberger2015u}
\bibfield{author}{\bibinfo{person}{Olaf Ronneberger}, \bibinfo{person}{Philipp Fischer}, {and} \bibinfo{person}{Thomas Brox}.} \bibinfo{year}{2015}\natexlab{}.
\newblock \showarticletitle{{U-Net}: Convolutional networks for biomedical image segmentation}. In \bibinfo{booktitle}{\emph{Medical Image Computing and Computer-Assisted Intervention--MICCAI 2015: 18th International Conference, Munich, Germany, October 5--9, 2015, Proceedings, Part III 18}}. \bibinfo{publisher}{Springer International Publishing}, \bibinfo{pages}{234--241}.
\newblock


\bibitem[Ryali et~al\mbox{.}(2023)]%
        {ryali2023hiera}
\bibfield{author}{\bibinfo{person}{Chaitanya Ryali} {et~al\mbox{.}}} \bibinfo{year}{2023}\natexlab{}.
\newblock \showarticletitle{Hiera: A hierarchical vision transformer without the bells-and-whistles}. In \bibinfo{booktitle}{\emph{International Conference on Machine Learning (ICML)}}. PMLR, \bibinfo{pages}{29441--29454}.
\newblock


\bibitem[Scimone et~al\mbox{.}(2021)]%
        {scimone2021assessment}
\bibfield{author}{\bibinfo{person}{Mark~T Scimone}, \bibinfo{person}{Savitri Krishnamurthy}, \bibinfo{person}{Gopi Maguluri}, \bibinfo{person}{Dorin Preda}, \bibinfo{person}{Jesung Park}, \bibinfo{person}{John Grimble}, \bibinfo{person}{Min Song}, \bibinfo{person}{Kechen Ban}, {and} \bibinfo{person}{Nicusor Iftimia}.} \bibinfo{year}{2021}\natexlab{}.
\newblock \showarticletitle{Assessment of breast cancer surgical margins with multimodal optical microscopy: A feasibility clinical study}.
\newblock \bibinfo{journal}{\emph{PLOS One}} \bibinfo{volume}{16}, \bibinfo{number}{2} (\bibinfo{year}{2021}), \bibinfo{pages}{e0245334}.
\newblock


\bibitem[Siegel et~al\mbox{.}(2023)]%
        {siegel2023cancer}
\bibfield{author}{\bibinfo{person}{Rebecca~L Siegel}, \bibinfo{person}{Kimberly~D Miller}, \bibinfo{person}{Nikita~Sandeep Wagle}, {and} \bibinfo{person}{Ahmedin Jemal}.} \bibinfo{year}{2023}\natexlab{}.
\newblock \showarticletitle{Cancer statistics, 2023}.
\newblock \bibinfo{journal}{\emph{Ca Cancer J Clin}} \bibinfo{volume}{73}, \bibinfo{number}{1} (\bibinfo{year}{2023}), \bibinfo{pages}{17--48}.
\newblock


\bibitem[To et~al\mbox{.}(2022)]%
        {to2022deep}
\bibfield{author}{\bibinfo{person}{Tyrell To}, \bibinfo{person}{Saba~Heidari Gheshlaghi}, {and} \bibinfo{person}{Dong~Hye Ye}.} \bibinfo{year}{2022}\natexlab{}.
\newblock \showarticletitle{Deep learning for breast cancer classification of deep ultraviolet fluorescence images toward intra-operative margin assessment}. In \bibinfo{booktitle}{\emph{2022 44th Annual International Conference of the IEEE Engineering in Medicine \& Biology Society (EMBC)}}. IEEE, \bibinfo{pages}{1891--1894}.
\newblock


\bibitem[Veluponnar et~al\mbox{.}(2023)]%
        {veluponnar2023toward}
\bibfield{author}{\bibinfo{person}{Dinusha Veluponnar}, \bibinfo{person}{Lisanne~L de Boer}, \bibinfo{person}{Freija Geldof}, \bibinfo{person}{Lynn-Jade~S Jong}, \bibinfo{person}{Marcos Da~Silva~Guimaraes}, \bibinfo{person}{Marie-Jeanne~TFD Vrancken~Peeters}, \bibinfo{person}{Frederieke van Duijnhoven}, \bibinfo{person}{Theo Ruers}, {and} \bibinfo{person}{Behdad Dashtbozorg}.} \bibinfo{year}{2023}\natexlab{}.
\newblock \showarticletitle{Toward intraoperative margin assessment using a deep learning-based approach for automatic tumor segmentation in breast lumpectomy ultrasound images}.
\newblock \bibinfo{journal}{\emph{Cancers}} \bibinfo{volume}{15}, \bibinfo{number}{6} (\bibinfo{year}{2023}), \bibinfo{pages}{1652}.
\newblock


\bibitem[Ward et~al\mbox{.}(2025)]%
        {ward2025automated}
\bibfield{author}{\bibinfo{person}{Tyler Ward}, \bibinfo{person}{Braxton McFarland}, \bibinfo{person}{Sahar Nozad}, \bibinfo{person}{Talal Arshad}, \bibinfo{person}{Hafsa Nebbache}, \bibinfo{person}{Jin Chen}, \bibinfo{person}{Xiaoqin Wang}, {and} \bibinfo{person}{Abdullah-Al-Zubaer Imran}.} \bibinfo{year}{2025}\natexlab{}.
\newblock \showarticletitle{Automated Intraoperative Lumpectomy Margin Detection using SAM-Incorporated Forward-Forward Contrastive Learning}. In \bibinfo{booktitle}{\emph{Medical Imaging with Deep Learning-Short Papers}}.
\newblock


\bibitem[Xie et~al\mbox{.}(2024)]%
        {xie2024masksam}
\bibfield{author}{\bibinfo{person}{Bin Xie}, \bibinfo{person}{Hao Tang}, \bibinfo{person}{Bin Duan}, \bibinfo{person}{Dawen Cai}, {and} \bibinfo{person}{Yan Yan}.} \bibinfo{year}{2024}\natexlab{}.
\newblock \showarticletitle{{MaskSAM}: Towards Auto-prompt SAM with Mask Classification for Medical Image Segmentation}.
\newblock \bibinfo{journal}{\emph{arXiv preprint arXiv:2403.14103}} (\bibinfo{year}{2024}).
\newblock


\bibitem[Xu et~al\mbox{.}(2023)]%
        {xu2023sppnet}
\bibfield{author}{\bibinfo{person}{Qing Xu}, \bibinfo{person}{Wenwei Kuang}, \bibinfo{person}{Zeyu Zhang}, \bibinfo{person}{Xueyao Bao}, \bibinfo{person}{Haoran Chen}, {and} \bibinfo{person}{Wenting Duan}.} \bibinfo{year}{2023}\natexlab{}.
\newblock \showarticletitle{{SPPNet}: A single-point prompt network for nuclei image segmentation}. In \bibinfo{booktitle}{\emph{International Workshop on Machine Learning in Medical Imaging (MLMI)}}. Springer, \bibinfo{pages}{227--236}.
\newblock


\bibitem[Ye et~al\mbox{.}(2024)]%
        {ye2024continual}
\bibfield{author}{\bibinfo{person}{Yiwen Ye}, \bibinfo{person}{Yutong Xie}, \bibinfo{person}{Jianpeng Zhang}, \bibinfo{person}{Ziyang Chen}, \bibinfo{person}{Qi Wu}, {and} \bibinfo{person}{Yong Xia}.} \bibinfo{year}{2024}\natexlab{}.
\newblock \showarticletitle{Continual self-supervised learning: Towards universal multi-modal medical data representation learning}. In \bibinfo{booktitle}{\emph{Proceedings of the IEEE/CVF Conference on Computer Vision and Pattern Recognition (CVPR)}}. \bibinfo{pages}{11114--11124}.
\newblock


\bibitem[Yushkevich et~al\mbox{.}(2016)]%
        {yushkevich2016itk}
\bibfield{author}{\bibinfo{person}{Paul~A Yushkevich}, \bibinfo{person}{Yang Gao}, {and} \bibinfo{person}{Guido Gerig}.} \bibinfo{year}{2016}\natexlab{}.
\newblock \showarticletitle{ITK-SNAP: An interactive tool for semi-automatic segmentation of multi-modality biomedical images}. In \bibinfo{booktitle}{\emph{2016 38th annual international conference of the IEEE engineering in medicine and biology society (EMBC)}}. IEEE, \bibinfo{pages}{3342--3345}.
\newblock


\bibitem[Zhang et~al\mbox{.}(2024)]%
        {zhang2024evf}
\bibfield{author}{\bibinfo{person}{Yuxuan Zhang}, \bibinfo{person}{Tianheng Cheng}, \bibinfo{person}{Rui Hu}, \bibinfo{person}{Lei Liu}, \bibinfo{person}{Heng Liu}, \bibinfo{person}{Longjin Ran}, \bibinfo{person}{Xiaoxin Chen}, \bibinfo{person}{Wenyu Liu}, {and} \bibinfo{person}{Xinggang Wang}.} \bibinfo{year}{2024}\natexlab{}.
\newblock \showarticletitle{{EVF-SAM}: Early vision-language fusion for text-prompted {Segment Anything Model}}.
\newblock \bibinfo{journal}{\emph{arXiv preprint arXiv:2406.20076}} (\bibinfo{year}{2024}).
\newblock


\bibitem[Zhou et~al\mbox{.}(2024)]%
        {zhou2024sam}
\bibfield{author}{\bibinfo{person}{Chunpeng Zhou}, \bibinfo{person}{Kangjie Ning}, \bibinfo{person}{Qianqian Shen}, \bibinfo{person}{Sheng Zhou}, \bibinfo{person}{Zhi Yu}, {and} \bibinfo{person}{Haishuai Wang}.} \bibinfo{year}{2024}\natexlab{}.
\newblock \showarticletitle{{SAM-SP}: Self-prompting makes sam great again}.
\newblock \bibinfo{journal}{\emph{arXiv preprint arXiv:2408.12364}} (\bibinfo{year}{2024}).
\newblock


\end{thebibliography}

\end{document}